%% file: main.tex
\documentclass[11pt]{article}

\usepackage[final]{latex/acl}

\usepackage{times}
\usepackage{latexsym}
\usepackage[T1]{fontenc}
\usepackage[utf8]{inputenc}
\usepackage{microtype}
\IfFileExists{inconsolata.sty}{\usepackage{inconsolata}}{}
\usepackage{graphicx}
\usepackage{hyperref}
\usepackage{url}
\usepackage{anyfontsize}

\input{math_commands.tex}

\usepackage{listings}
\usepackage{xspace}
\usepackage{multirow}
\usepackage{booktabs}
\usepackage{amsmath, amssymb}

\usepackage{algorithm}
\usepackage{algpseudocode}
\usepackage{caption}
\usepackage[breakable]{tcolorbox}
\usepackage{enumitem}
\usepackage{amsthm}

\usepackage[table]{xcolor}

\definecolor{takeawayBg}{HTML}{EEF6FF}
\definecolor{takeawayRule}{HTML}{2F6FD6}

\tcbset{
  prismcasebox/.style={
    breakable,
    colback=takeawayBg,
    colframe=takeawayBg,
    coltitle=takeawayRule,
    colbacktitle=takeawayBg,
    boxrule=0pt,
    arc=1.4mm,
    boxsep=0pt,
    left=2mm,
    right=2mm,
    top=1.4mm,
    bottom=1.4mm,
    toptitle=0mm,
    bottomtitle=0.8mm,
    fonttitle=\bfseries\small
  },
  prismsolutionbox/.style={
    breakable,
    colback=white,
    colframe=black!25,
    coltitle=black,
    colbacktitle=white,
    boxrule=0pt,
    leftrule=0.4pt,
    arc=0mm,
    outer arc=0mm,
    boxsep=0pt,
    left=1.4mm,
    right=1mm,
    top=0.9mm,
    bottom=0.9mm,
    toptitle=0mm,
    bottomtitle=0.6mm,
    fonttitle=\bfseries\footnotesize
  }
}

\usepackage{bm} 

\usepackage{wrapfig}

\newcommand{\vect}[1]{\mathbf{#1}} 
\newcommand{\mat}[1]{\mathbf{#1}}

\newcommand{\ours}{\texttt{PRISM}\xspace}

\input{math_commands.tex}
\title{\raisebox{-0.22em}{\includegraphics[height=1.35em]{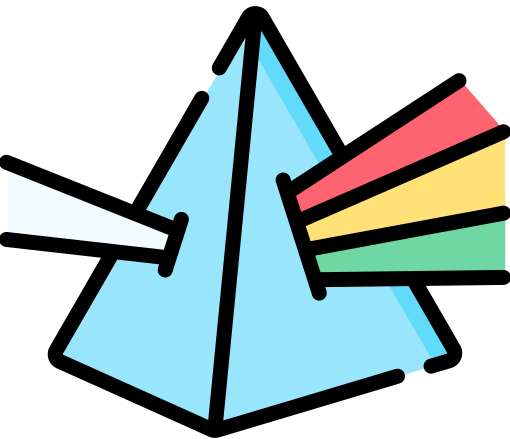}}~\ours: Promoting Sampling Diversity of Large Language Models with Implicit Stochastic Modulation}

\author{
\textbf{Qixin Xu\textsuperscript{1\ *}}\quad
\textbf{Zihan Zhu\textsuperscript{1}}\quad
\textbf{Xiaojiang Liu\textsuperscript{2}}\quad
\textbf{Bohan Lyu\textsuperscript{1\ *\ †}}
\\
\textsuperscript{1}{Tsinghua University.}
\quad
\textsuperscript{2}{Apple.}
\\
{\tt xqx23@mails.tsinghua.edu.cn, lyubh22@gmail.com}
\\[1ex]
\textsuperscript{*}Equal contribution
\hspace{2em}
\textsuperscript{\dag}Corresponding author
}

\begin{document}

\maketitle

\begin{abstract}
Large language models (LLMs) exhibit \emph{mode collapse}, where their output distribution concentrates on familiar semantic modes. When diversity disappears, repeated sampling becomes semantic repetition rather than discovery, making rare but stronger solutions unlikely to surface and leaving self-evolution and continual learning difficult to sustain. Most remedies intervene only at inference time, shaping samples without making diversity a model capability.
We propose \ours, a training-time framework that internalizes diversity within the model. \ours combines (i) \emph{submodular curation} that pairs each prompt with diverse-yet-correct multi-teacher solutions, and (ii) \emph{stochastic embedding modulation}, which turns special-token perturbations into implicit signals for sampling distinct outputs. A theoretical analysis establishes the stability of this mechanism.
On WritingPrompts and BookMIA, \ours moves the quality-diversity frontier outward, achieving the best generation quality in all model-benchmark settings while maintaining the strongest or near-strongest semantic diversity across two open-weight model families. It also improves sampling efficiency on math and code benchmarks. Beyond inference, \ours yields a stronger foundation for reinforcement learning, sustaining diverse rollouts and mitigating the diversity collapse commonly observed during RL training.
\end{abstract}

\input{sections/1_intro}
\input{sections/3_method}

\input{sections/4_experiment}
\input{sections/2_relatedwork}
\input{sections/5_conclusion}

\section*{Limitations}

PRISM depends on a multi-solution candidate pool. For math and code we draw on existing corpora, but for creative writing we constructed the pool by repeated sampling from eight teacher models, a cost that recurs in any domain without off-the-shelf multi-solution data. The noise scale is task-dependent (sigma = 0.005 for writing, 0.01 for reasoning; Appendix E), requiring a selection pass per task family. Our analysis establishes stability but not diversity at the objective level: the expected objective admits a noise-invariant optimum, and the routing behavior we observe is a property of finite-step optimization, characterized empirically rather than derived; making the noise informative by construction is an open direction. Finally, evaluation relies on LLM judges and a single embedding model, and our experiments cover only 7B to 8B models.

\bibliography{main}

\appendix

\input{sections/6_appendix}

\end{document}

%% file: math_commands.tex
\usepackage{amsmath,amsfonts,bm}

\def\eqref#1{equation~\ref{#1}}

\def\1{\bm{1}}

\DeclareMathAlphabet{\mathsfit}{\encodingdefault}{\sfdefault}{m}{sl}
\SetMathAlphabet{\mathsfit}{bold}{\encodingdefault}{\sfdefault}{bx}{n}



%% file: sections/1_intro.tex
\section{Introduction}

Post-trained large language models (LLMs) exhibit \emph{mode collapse}, where their output distribution concentrates on a small set of familiar semantic modes~\citep{padmakumar2024doeswritinglanguagemodels, west2025basemodelsbeataligned, zhang2025verbalizedsamplingmitigatemode, lanchantin2025diversepreferenceoptimization}. This collapse weakens the model's ability to use sampling as exploration: repeated samples revisit the same trajectories, making rare but stronger outputs harder to discover and limiting the rollout diversity needed for self-improvement. The effect appears in both open-ended generation, where outputs become homogeneous~\citep{lu2025aihumanityssalieriquantifying, zhang2025verbalizedsamplingmitigatemode, jiang2025artificialhivemind}, and verifiable reasoning, where near-identical solution paths reduce the gains from scaled test-time compute~\citep{snell2024scaling, muennighoff2025s1, hooper2025ets, ju2025reasoningpathdivergence, lyu2025adapting, wang2024planningnaturallanguageimproves, song2025outcomebasedexploration}. Because diversity loss can be shaped by post-training data and learned weights, not only decoding choices~\citep{karouzos2026wheredoesoutputdiversitycollapse}, reducing mode collapse is crucial for preserving a model's capacity to explore, discover, and self-evolve.

\begin{figure*}[!t]
    \centering
    \includegraphics[width=\linewidth]{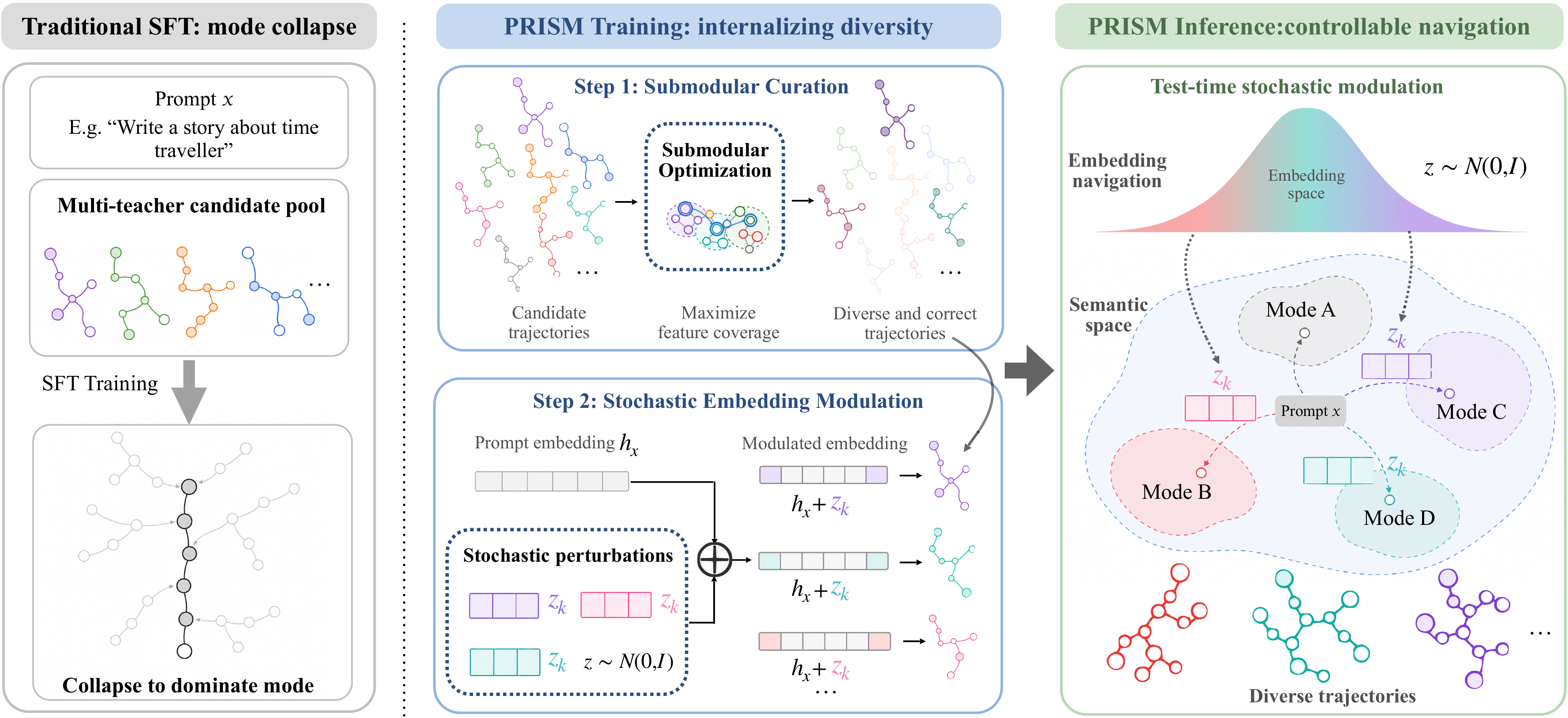}
    \vspace{-3mm}
    \caption{Overview of \ours. \textbf{Left:} Traditional SFT collapses onto a single dominant trajectory.
\textbf{Middle:} \ours{} internalizes diversity through submodular curation of diverse-yet-correct trajectories and stochastic embedding modulation that perturbs the prompt embedding with Gaussian noise.
\textbf{Right:} At inference, different noise samples steer the model toward distinct modes, producing diverse trajectories from the same prompt.}
    \label{fig:rep}
\end{figure*}

Existing methods for improving LLM output diversity fall into several families, but they rarely make diversity something the model can be steered to produce on demand. \emph{Inference-time methods}, including token-level decoding strategies~\citep{vijayakumar2018diversebeamsearchdecoding, ippolito2019comparison, meister2023locally}, prompt-based interventions~\citep{naik2023diversity, zhang2025verbalizedsamplingmitigatemode, wong2025simplestrat, ruan2025g2, santos2025diverseprompts}, search-based methods~\citep{mouret2015illuminating, romera2024funsearch, novikov2025alphaevolve, he2026diffcodegen}, and latent or decoding-space alternatives~\citep{geiping2025scalingtesttimecomputelatent, zhang2025softthinkingunlockingreasoning, chen2025flaminghotinitiationregularexecution}, try to force variation at generation time. Because the model itself has not learned when and how to vary its answers, these methods can stop at surface-level changes or spend extra tokens for each prompt. \emph{Training-time noise methods}, in contrast, do modify the model itself, but typically inject noise into embeddings as a generic regularizer~\citep{jain2023neftunenoisyembeddingsimprove, yadav2023symnoise, Zhu2020FreeLB:}, improving instruction-following or robustness without making diversity controllable. Recent data-centric and multi-solution training methods improve coverage, especially in reasoning and code, but they largely decide which traces to train on or how to label modes~\citep{wang2024dolphcoder, ju2025reasoningpathdivergence, just2025signalsteps, jia2026traininglargelanguagemodelsreasonparallel}. The remaining gap is a model-level mechanism that lets the same prompt produce different valid outputs through an internal control signal.

To this end, we propose \ours, a training-time framework that teaches LLMs to use a small internal signal to choose among multiple valid outputs for the same prompt. \ours has two coupled components: one defines \emph{what} diverse outputs the model should learn, and the other gives the model a way to move among them. First, we build specialized training datasets via submodular curation, in which each prompt is paired with multiple diverse-yet-correct solutions drawn from a multi-teacher candidate pool; this defines the \emph{target space} of valid output trajectories the model should be able to produce. Second, we inject Gaussian noise into a small set of special-token embeddings during both training and inference; under the curated supervision provided by the first component, different noise samples become \emph{learned navigation signals} that steer the model toward distinct trajectories within this target space, rather than acting as generic regularization. We provide a theoretical analysis showing that this noise injection is provably stable and can be interpreted as a regularizer on the Hessian of the loss with respect to embeddings.


We center our evaluation on open-ended creative writing, where diversity must be semantic rather than merely lexical. On WritingPrompts and BookMIA across two open-weight model families, \ours achieves the best generation quality in all model-benchmark settings and the strongest or near-strongest semantic diversity among competitive inference-time and training-time baselines. These results shift the quality-diversity frontier outward rather than trading one axis for the other. We further evaluate \ours on verifiable math and code reasoning, where it improves sampling efficiency by discovering more correct solutions under equivalent inference budgets.

For reinforcement learning (RL) post-training, diversity determines what behaviors the policy can still explore: once rollouts collapse, reward feedback comes from repeated variants of the same solution and policy improvement stalls~\citep{cui2025entropymechanismreinforcementlearning, cui2025process, wang2025emergenthierarchicalreasoningllms, xu2025cogdoc}. By exposing multiple correct modes through a controllable sampling interface, \ours lets RL train on a wider range of rollouts; in math and code reasoning, this preserves diversity during training and avoids the sampling collapse observed with standard SFT initialization.

In conclusion, our main contributions are:
\begin{itemize}
    \item We introduce \ours, a training-time framework that pairs per-prompt submodular curation with stochastic special-token embeddings to make diversity controllable at inference.
    \item We show that \ours improves the quality-diversity frontier on creative writing tasks on two open-weight models.
    \item We demonstrate stronger sampling-based exploration on eight math and code benchmarks, discovering more correct solutions than SFT and prior diversity-enhancing baselines under the same inference budgets.
    \item We show that diversity supports policy improvement: \ours checkpoints preserve exploratory rollouts during RL and help the policy continue discovering correct programs.
\end{itemize}

%% file: sections/3_method.tex
\section{Methodology}

Our methodology aims to enable LLMs to generate diverse yet correct reasoning trajectories. We first construct diverse training datasets using submodular optimization to sample varied reasoning paths (Section~\ref{sec:diverse_sampling_submodular}), then train and do inference with the model with noisy special-token embeddings (Section~\ref{sec:nem}). We provide analysis on our training target to understand its impact (Section~\ref{sec:training_target_analysis}).

\subsection{Diverse Reasoning Trajectory Sampling via Submodular Optimization}
\label{sec:diverse_sampling_submodular}

Following prior work on diversity as feature coverage in summarization and data selection 
\citep{Filatova2004,Takamura2009,Lin2011}, 
we define the diversity of a subset $A \subseteq S$ as the number of distinct reasoning features it covers:
\begin{equation}
    f(A) = \left| \bigcup_{s_a \in A} F(s_a) \right|,
    \label{eq:submodular_objective}
\end{equation}
where $F(s_a)$ denotes the set of discrete reasoning features of trajectory $s_a$. Feature construction is per-domain: for creative writing, we embed candidate completions with SBERT and cluster them, using cluster IDs as the discrete feature set; for math, features are n-grams over solution tokens; for code, n-grams together with AST subtrees. Appendix~\ref{app:feature_extraction_formal} details each construction.

To maximize this diversity under a budget constraint, we adopt a two-stage framework. 
First, for each input, we collect a candidate set of $m$ trajectories, $S = \{s_1, \dots, s_m\}$. 
Second, we employ a submodular subset optimization strategy~\citep{krause2014submodular} 
to extract a fixed-size subset $A \subseteq S$ of $n$ trajectories that maximizes $f(A)$.

We use a greedy algorithm to maximize $f(A)$ subject to $|A|=n$. 
Starting with an empty set $A_0 = \emptyset$, at each step $t$ ($t=0, \dots, n-1$), 
we select trajectory $s_{t+1}$ that offers the largest marginal gain:
\begin{equation}
    s_{t+1} = \underset{s_i \in S \setminus A_t}{\arg\max} \big( f(A_t \cup \{s_i\}) - f(A_t) \big).
    \label{eq:greedy_selection_general}
\end{equation}
Let $U_t = \bigcup_{s_a \in A_t} F(s_a)$ denote the features already covered. 
The criterion simplifies to:
\begin{equation}
    s_{t+1} = \underset{s_i \in S \setminus A_t}{\arg\max} \; \big| F(s_i) \setminus U_t \big|.
    \label{eq:greedy_selection_specific}
\end{equation}
This algorithm guarantees a $1 - 1/e$ approximation to the optimal feature coverage achievable with $n$ samples~\citep{nemhauser1978analysis}. Appendix~\ref{app:submodular_details_formal} further explains it.
We further validate the selected training data with human annotations on 32 code and math queries. As shown in Table~\ref{tab:formalizer}, submodular selection produces more queries with at least three distinct solutions than random sampling; annotation details are in Appendix~\ref{appx: human check}.

\begin{table}[t]
\centering
\small
\caption{Human-evaluated solution diversity in training data with different sampling methods.}
\resizebox{\columnwidth}{!}{\begin{tabular}{lccc} 
\toprule
\textbf{Datasets} & $\ge 3$ & $2$ & $1$ \\
\midrule
Code & $34.4$ & $37.5$ & $28.1$ \\
\hspace{1mm} w/o submodular optimization & $25.0$ & $37.5$ & $37.5$ \\
\midrule
Math & $42.8$ & $39.3$ & $17.9$ \\
\hspace{1mm} w/o submodular optimization & $28.1$ & $37.5$ & $34.4$ \\
\bottomrule
\end{tabular}}
\label{tab:formalizer}
\end{table}







\subsection{Noisy Special-token Embeddings as Implicit Diversity Signals}
\label{sec:nem}

While traditional SFT learns a direct mapping between inputs and outputs, it does not account for the need to generate diverse yet correct outputs for the same input. To foster diversity-controllable sampling, we hope to control the models to produce different valid reasoning paths.

To achieve this, we introduce Gaussian noise perturbations to the embeddings of special input tokens during both training and inference. Let $x = (x_1, \dots, x_L)$ be the token IDs of an input sequence. The standard sequence of embedding vectors for $x$ is $\mathbf{E}_x = (\mathbf{e}_{x_1}, \dots, \mathbf{e}_{x_L})$. We define a corresponding noise sequence $\mathbf{N}_x = (\mathbf{n}_{x_1}, \dots, \mathbf{n}_{x_L})$. For each position $j$, the noise vector $\mathbf{n}_{x_j}$ is sampled from $\mathcal{N}(\mathbf{0}, \sigma^2 \mathbf{I})$ if $x_j$ is a special token, and $\mathbf{n}_{x_j} = \mathbf{0}$ if $x_j$ is not. This targeted noise injection perturbs the input representation at specific special token positions, thereby stimulating diverse model outputs while keeping the core semantic information of the input sequence $x$ intact.
The effective input embedding sequence to the model, $\mathbf{E}'_x$, is the element-wise sum of the two parts:
\begin{equation} \label{eq:noisy_embedding_sequence_sum}
\mathbf{E}'_x = \mathbf{E}_x + \mathbf{N}_x = (\mathbf{e}_{x_1} + \mathbf{n}_{x_1}, \dots, \mathbf{e}_{x_L} + \mathbf{n}_{x_L}).
\end{equation}




The training objective is
\begin{equation}
\mathcal{J}(\theta) = \mathbb{E}_{x,\, y \sim A(x),\, \mathbf{N}_x}
\left[ -\log p_\theta(y \mid \mathbf{E}'_x) \right]
\label{eq:main_loss_in_lm_paragraph}
\end{equation}
where $A(x)$ is the curated answer set from
Section~\ref{sec:diverse_sampling_submodular} and $\mathbf{N}_x$ is
freshly sampled per forward pass. Because the same $x$ is paired with multiple $y \in A(x)$ under
different $\mathbf{N}_x$ over the course of training,
noise-conditioned solutions that commit to different valid answers
under different draws attain the same expected loss as any
noise-invariant one, and training settles on the former: the model
learns to treat the noise as an implicit selector among valid
answers. At inference, varying $\mathbf{N}_x$ then yields varying $y$.

\subsection{Training Target Analysis}
\label{sec:training_target_analysis} 

We conduct theoretical analyses to answer two questions: (1) Does injecting noise into special-token embeddings destabilize the language model? (2) How does our training formulation facilitate controllable diversity at inference time?

\paragraph{Embeddings.}
For the embeddings, the introduction of Gaussian noise acts as a form of regularization. Formally, let the vocabulary size be $V$ and the embedding dimension be $D$. The embedding matrix is denoted as $\mathbf{E} \in \mathbb{R}^{V \times D}$. $\mathbf{E}'$ is the embedding added with randomly sampled Gaussian Noise. Analyzing the expected loss function $\mathbb{E}_{\mathbf{n}} [ L (\mathbf{E}') ]$ reveals that it introduces an regularization term:
\begin{equation}
\mathbb{E}_{\mathbf{n}} [ L (\mathbf{E}') ] = L(\mathbf{E}) + \frac{\sigma^2}{2} \sum_{i \in \mathcal{S}} \text{tr}(\mathbf{H}_i) + O(\sigma^3),
\end{equation}
where $\mathbf{H}_i$ is the Hessian of the loss function with respect to the embedding $\mathbf{e}_i$. This term, proportional to the trace of the Hessian, is a style of regularization~\citep{bishop1995training} by penalizing high-curvature regions of the loss landscape. This justifies the noise injection as a principled regularizer, contributing a well-behaved smoothing term to the expected loss. See Appendix~\ref{sec:proof_reg} for a detailed derivation. 

\paragraph{Language Model.}


We can decompose the per-instance log-probability term in
Equation~\ref{eq:main_loss_in_lm_paragraph} as (noting that
$y \in A(x)$ by construction):
{\small
\begin{equation}
\begin{aligned}
    &p_\theta(y \mid \mathbf{E}'_x) = \\
    &\qquad p_\theta\bigl(Y = y \mid Y \in A(x), \mathbf{E}'_x\bigr)
    \cdot p_\theta\bigl(Y \in A(x) \mid \mathbf{E}'_x\bigr).
\end{aligned}
\label{eq:prob_decomposition}
\end{equation}
}
Taking the negative logarithm, the per-instance loss decomposes
into a component ensuring general answer correctness and another
that promotes the noise-conditioned selection of specific diverse
answers:
{\small
\begin{equation}
\begin{aligned}
    &-\log p_\theta(y \mid \mathbf{E}'_x) =
    -\underbrace{\log\Bigl(\sum\nolimits_{y' \in A(x)} p_\theta(y' \mid \mathbf{E}'_x)\Bigr)}_{\text{Semantic Correctness}} \\
    &\qquad -\underbrace{\log p_\theta\bigl(Y = y \mid Y \in A(x), \mathbf{E}'_x\bigr)}_{\text{Noise-Conditioned Specificity}}.
\end{aligned}
\label{eq:loss_decomposition_in_main_text}
\end{equation}
}


The semantic correctness term ensures that the model consistently generates outputs from the set of correct solutions. The noise-conditioned specificity term is minimized by solutions that use the injected noise as a control signal to select a particular trajectory from the correct set, and our ablations in Section 3.1 confirm training converges to such a solution. At test time, by varying the noise injected, we can flexibly generate diverse valid outputs for the same input, thus achieving test-time scaling. A detailed explanation of this part is provided in Appendix~\ref{app:loss_decomposition_detail}.

%% file: sections/4_experiment.tex
\input{sections/table_cw}

\input{sections/tab_sft}

\section{Experiments}
\label{sec:exp}

We evaluate \ours on two complementary task families: \textbf{open-ended creative writing} (Section~\ref{sec:cw}), where quality is assessed jointly with diversity, and \textbf{verifiable reasoning} on math and code (Section~\ref{sec:reasoning}), where sampling efficiency is measured by Pass@N.
We then evaluate the \ours as a beneficial foundation for RL (Section~\ref{sec:rl}).

\subsection{Creative Writing}
\label{sec:cw}

\paragraph{Setup.}
We use two long-form benchmarks: \textbf{WritingPrompts}~\citep{fan2018hierarchicalneuralstorygeneration}, where each Reddit-style prompt asks the model to write a full short story, and \textbf{BookMIA}~\citep{shi2023detecting}, where the model continues a passage from the opening sentences of a book. Both yield outputs of 800--2000 tokens, leaving sufficient room for the noise-induced diversity signal to accumulate across plot decisions. For each benchmark we sample 100 test prompts and generate $N=30$ responses per prompt with target length 800 words. We evaluate two open-weight instruction-tuned models from different families, \texttt{Qwen2.5-7B-Instruct} and \texttt{Llama-3.1-8B-Instruct}, and compare \ours against five baselines spanning inference-time and training-time: Vanilla sampling, standard SFT, NEFTune~\citep{jain2023neftunenoisyembeddingsimprove}, VS-Standard~\citep{zhang2025verbalizedsamplingmitigatemode}, and FIRE~\citep{chen2025flaminghotinitiationregularexecution}. Details and hyperparameters are in appendix \ref{appx: baseline}.

\begin{figure*}[!t]
    \centering
    \includegraphics[width=\linewidth]{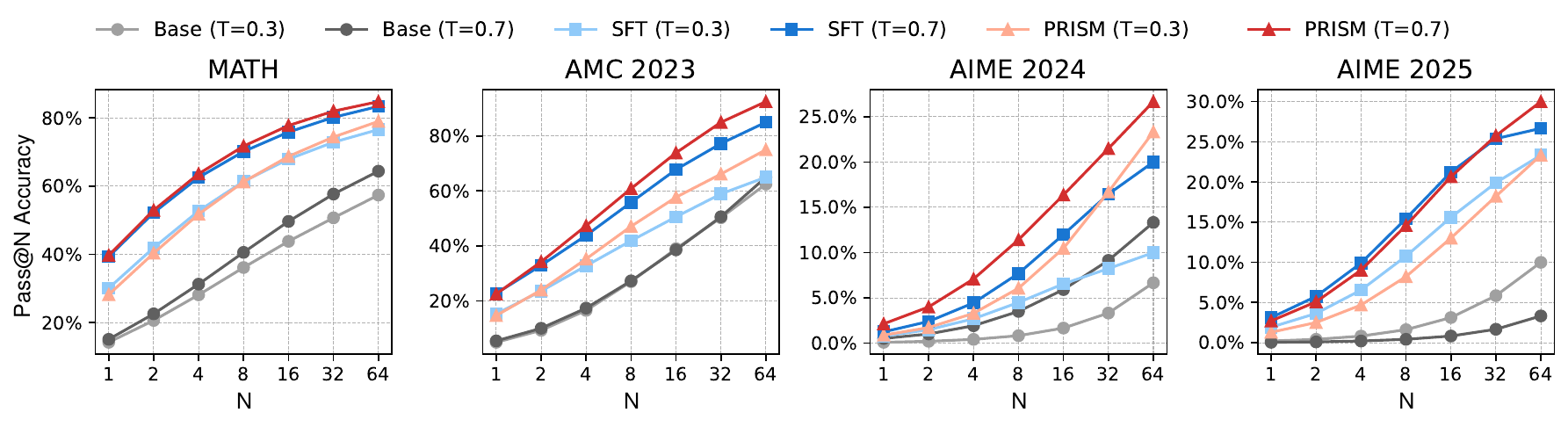}
    \vspace{-20pt}
    \caption{Pass@N performance of \ours, SFT, and Base models under different temperatures. \ours's gain is consistent across temperatures.}
    \label{fig:mathcomp}
\end{figure*}

\paragraph{Training Data.}
Training prompts are drawn from the WritingPrompts training split~\citep{fan2018hierarchicalneuralstorygeneration} and the BookMIA training split~\citep{shi2023detecting}, yielding 10000 prompts in total; the benchmark test prompts used for evaluation are held out throughout. For each training prompt, we follow the submodular selection pipeline detailed in (Section~\ref{sec:reasoning}), details are listed in appendix \ref{appx: training data}.

\paragraph{Evaluation.}
Following~\citet{zhang2025verbalizedsamplingmitigatemode}, we measure \textbf{diversity} as $1-\bar{s}$ (\%), where $\bar{s}$ is the mean pairwise cosine similarity of within-prompt response embeddings under OpenAI \texttt{text-embedding-3-small} (negative similarities clipped to $0$). \textbf{Quality} is the Creative Writing v3 rubric~\citep{creative-writing-bench-v3} score (0--100) judged by Claude-3.7-Sonnet and GPT-4o. Higher is better on both axes.

\paragraph{Results.}
Table~\ref{tab:cw_results} reports diversity and quality across both benchmarks and both model families. \ours achieves the highest quality in all four (model, benchmark) cells and strictly Pareto-dominates Vanilla, SFT, NEFTune, and FIRE in every cell. GEM is the strongest diversity baseline on Qwen, exceeding \ours by at most 1.3 points of diversity at a cost of 1.6 to 6.4 points of quality; on Llama, \ours is higher than GEM on both axes in both benchmarks. VS-Standard similarly exceeds \ours by at most 1.8 diversity points while giving up 5 to 8 points of quality. Across all four cells, \ours thus shifts the quality-diversity Pareto frontier outward rather than trading along it.

\paragraph{Ablations.} We evaluate four variants: (i) w/o Noise (curated data only, no noise), (ii) w/o Div (noise on both sides but randomly sampled training data), (iii) w/o InN (full training, no noise at inference), and (iv) SFT w/ InN (the SFT checkpoint with the same noise injected only at inference), which completes the 2×2 factorial over training-time and inference-time noise. The four comparisons attribute the gains component by component. Curation defines the modes: replacing curated data with random sampling (w/o Div) drops diversity by 3.6 to 4.6 points with noise fully in place. Curated data alone does not realize the gains: training on the same data without noise (w/o Noise) loses 6 to 9 points of quality relative to \ours. Training-time noise is what makes the perturbation usable: injected into a checkpoint never trained under it (SFT w/ InN), the same noise raises diversity but costs 3 to 6 points of quality, while under the noise-trained checkpoint it raises both. Finally, the mechanism requires inference-time activation: with identical training, w/o InN forfeits both the remaining diversity and a substantial quality margin (45.4 to 51.0 on Qwen WritingPrompts), showing that the noise functions as an inference-time conditioning signal rather than a training regularizer whose effect would already be in the weights.

\subsection{Math and Code Reasoning}
\label{sec:reasoning}

\paragraph{Setup.}
We evaluate on eight reasoning benchmarks: four code (HumanEval, MBPP, HumanEval+, MBPP+~\citep{chen2021evaluatinglargelanguagemodels, austin2021programsynthesislargelanguage, liu2023is}) and four math (MATH-500~\citep{hendrycks2021measuring, lightman2023letsverifystepstep}, AMC23, AIME 24/25~\citep{aime24, aime25}). The experiments are built on \texttt{Llama-3.1-8B-Base} with a warm-up SFT on Ultra-Chat~\citep{ding2023enhancing} and reasoning data distilled from \texttt{Llama-3.1-8B-Instruct}.

\paragraph{Training Data.}
For \textbf{math}, we sample 10000 problems from A-m-team verified data~\citep{tian2025deepdistillenhancingllmreasoning} with mid-range cross-model accuracy ($33\%$--$67\%$) and use n-gram submodular optimization to select 4 diverse answers per problem. For \textbf{code}, we draw from four sources: A-m-team distilled solutions, LeetCode problems, OpenCoder SFT-Stage-2 educational-instruct~\citep{Huang2024OpenCoderTO} solutions, and a broader OpenCoder mix for general coding ability. Code submodular selection uses both n-gram and AST features. Baseline datasets randomly sample 4 solutions per problem to control for token count. Full data preparation details are in appendix \ref{appx: training data}.

\paragraph{Evaluation.} We report Pass@N using the standard unbiased estimator~\citep{chen2021evaluatinglargelanguagemodels}. Given $M$ generated samples with $c$ correct ones, Pass@$k$ is estimated as $1 - \binom{M-c}{k}/\binom{M}{k}$ and averaged across all problems in each dataset.

\paragraph{Results.}
Table~\ref{tab:combined_results} shows \ours attains the highest Pass@N on seven of the eight benchmarks, with AIME 2025 tied with SFT and NEFTune higher on MBPP+. Against Base, \ours lifts AIME 2024 from 6.67 to 23.33 on a benchmark where mode collapse is most punitive. Against SFT, which already absorbs the benefit of supervised distillation, \ours still delivers consistent further gains, including +13.33 on AIME 2024 and +10.0 on AMC 23. NEFTune, which injects noise only during training as a regularizer, also improves clearly over SFT, consistent with the robustness effect of training-time noise; the remaining gap to \ours, largest on AIME 2024 where trajectory diversity matters most, is what inference-time conditioning adds. Figure~\ref{fig:mathcomp} further shows these gains hold across temperatures, confirming that \ours's mechanism is complementary to, rather than redundant with, temperature scaling.

\paragraph{Ablations.} We evaluate the w/o Noise, w/o Div, and
w/o InN variants on the reasoning benchmarks. Each component alone is insufficient: w/o Noise improves over SFT on math but is flat on code, and w/o Div lifts math further yet underperforms SFT on several code benchmarks. More importantly, inference-time noise is what activates the learned diversity signal: w/o InN consistently underperforms full \ours despite sharing identical training (AIME 2024 pass@64: 16.67 vs. 23.33), confirming that the noise vector functions as a learned navigation signal exercised at test time rather than a training-time regularizer.

\input{sections/tab_rl}

\subsection{Reinforcement Learning}
\label{sec:rl}

\paragraph{Setup.}
RL training is particularly vulnerable to rollout-diversity collapse, where decreasing policy entropy stalls improvement~\citep{cui2025entropymechanismreinforcementlearning, lin2025goedel}. We test whether \ours's training-time diversity carries into RL by initializing veRL~\citep{sheng2024hybridflow} from two checkpoints, \texttt{SFT}, and full \texttt{\ours}, and training on SimpleRL-Zoo~\citep{zeng2025simplerlzooinvestigatingtamingzero} for math and Acecoder-V2~\citep{zeng2025acecoderacingcoderrl} for code. The full \ours run additionally injects noise during RL rollouts. Details of RL training are in Appendix~\ref{appx:rl}.

\paragraph{Results.}
Table~\ref{tab:combined_results_rl} compares pre- and post-RL pass@64
across math and code; AIME averages AIME 2024 and AIME 2025. Two
findings stand out. First, \texttt{\ours} provides a uniformly stronger
RL initialization: both pre-RL and post-RL pass@64 are higher than
\texttt{SFT} in every cell, indicating that the diversity inherited
from \texttt{\ours} persists as a more productive search space
throughout RL. Second, math and code respond asymmetrically. On math,
both \texttt{SFT} and \texttt{\ours} improve after RL. On code,
\texttt{SFT} loses pass@64 across all four settings (HumanEval/+ at
both temperatures), whereas \texttt{\ours} improves on three of four
code cells and drops only marginally on HumanEval+ at $T{=}0.3$.
\texttt{\ours} thus appears to prevent the high-pass collapse commonly
observed during RL fine-tuning. We attribute the math--code asymmetry
to the distinct distributional shifts each task imposes between SFT
and RL training data.

\section{Analyses}

\subsection{Other Diversity Enhancing Baselines}
We also compare our methods with other methods that focus on reasoning diversity on code tasks.

\paragraph{Baselines.}
(1) Diverse Beam Search~(DBS)~\citep{vijayakumar2018diversebeamsearchdecoding}: A beam search-based method that optimizes a diversity-augmented objective. We applied DBS to both \texttt{Llama-3.1-8B-Base} with warm-up and \texttt{Llama-3.1-8B-Instruct}. (2) GEM~\citep{li2025preservingdiversitysupervisedfinetuning}: GEM is an SFT algorithm that prevents over-memorization and promotes diversity in the model's outputs. We applied GEM to \texttt{Llama-3.1-8B-Base} with randomly-sampled training data and default training parameters.

\paragraph{Results.}
As shown in Table~\ref{tab:code_results1}, our analysis reveals several key findings. (1) DBS consistently degrades performance on both base and instruct models. It results in lower scores across all Pass@1 and Pass@N metrics compared to its respective original model. (2) \ours demonstrates superior performance especially at higher sampling budgets. While GEM performs well on HumanEval, our method achieves the highest Pass@64 scores on the other three benchmarks, outperforming both GEM and the extensively trained \texttt{Llama-3.1-8B-Instruct} model.

\begin{table}[!ht]
\centering
\caption{Pass@64 performance on code benchmarks. \ours demonstrates superior sampling efficiency compared with both GEM and DBS.}
\label{tab:code_results1}
\resizebox{\columnwidth}{!}{
\begin{tabular}{lcccc}
\toprule
Method & HumanEval & HumanEval+ & MBPP & MBPP+ \\
\midrule
Base & $81.14$ & $72.89$ & $75.42$ & $69.99$ \\
\; w/ DBS & $62.58$ & $56.10$ & $66.40$ & $61.64$ \\
Instruct & $\underline{88.49}$ & $\underline{83.27}$ & $\underline{86.05}$ & $\underline{76.42}$ \\
\; w/ DBS & $79.75$ & $73.17$ & $85.45$ & $74.87$ \\
GEM & $\mathbf{89.57}$ & $82.32$ & $84.13$ & $75.66$ \\
\rowcolor{blue!10}
\ours & $87.81$ & $\mathbf{83.82}$ & $\mathbf{86.11}$ & $\mathbf{77.26}$ \\ \bottomrule
\end{tabular}}
\end{table}

\subsection{Sourcing the Effectiveness of \ours}
To elucidate how \ours improves Pass@N performance in reasoning task, we conduct analyses on code tasks. We annotate the algorithms utilized in the solutions generated by \ours and SFT models: we firstly categorize all possible solution algorithms into 39 classes and employ the Qwen3-8B model to automatically label each generated solution. The details for labeling are in Appendix~\ref{appx: llmasjudge}.

\begin{figure}[!ht]
    \centering
    \includegraphics[width=\columnwidth]{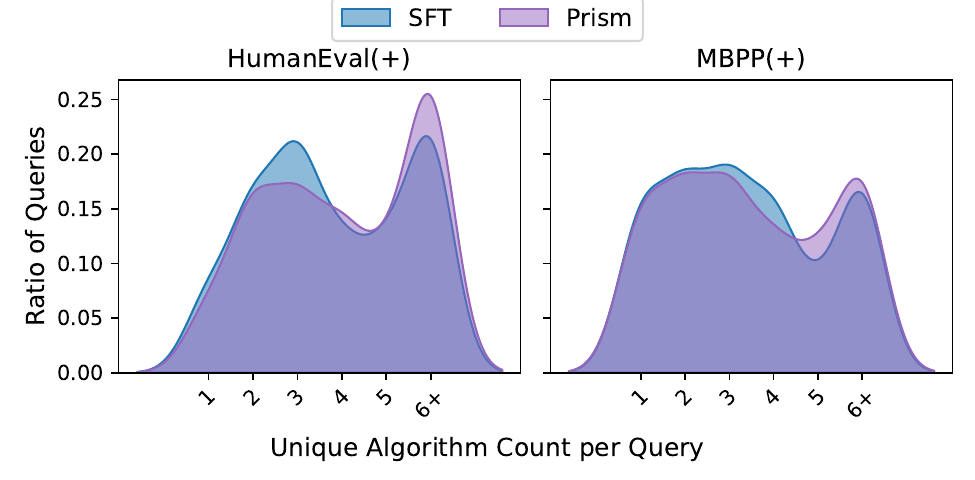}
    \caption{Distribution of unique algorithm counts per query for SFT and \ours. \ours{}'s shifts toward higher counts, indicating greater algorithmic diversity.}
    \label{fig:combined-performance}
\end{figure}

\paragraph{\ours Enhances Algorithmic Diversity.} For each query, we compute the change in both unique algorithm count and Pass@N score between \ours and SFT. As shown in Figure~\ref{fig:combined-performance}, \ours consistently shifts the distribution towards a higher count of unique algorithms per query, which indicates a greater algorithm-level diversity in its outputs compared to SFT.

\begin{figure}[!ht]
    \centering
    \includegraphics[width=\columnwidth]{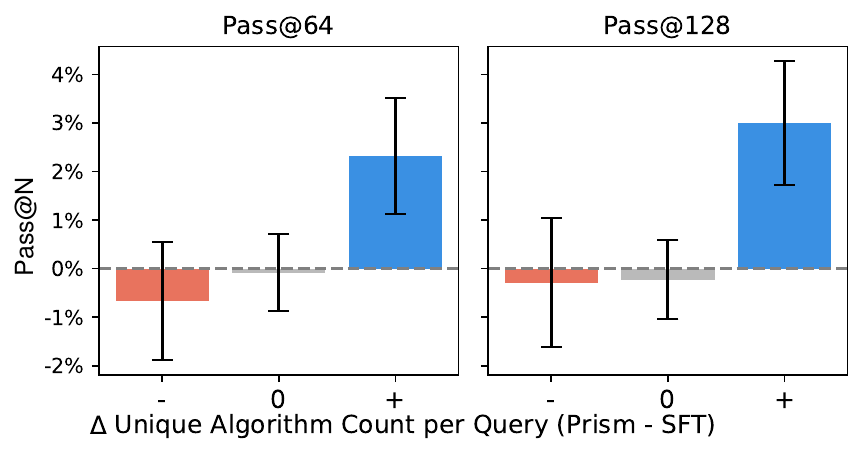}
    \caption{Change in Pass@N versus change in unique algorithm count
    per query (both \ours{} minus SFT), grouped into decrease (-),
    unchanged (0), and increase (+). Error bars show standard error
    of the mean.}
    \label{fig:reg}
\end{figure}

\paragraph{Algorithmic Diversity Drives Test-Time Scaling.} We then statistically examine the link between this enhanced diversity and performance gains. For each query, we compute the delta in both the unique algorithm count and the Pass@N score between the \ours and SFT models. As illustrated in Figure~\ref{fig:reg}, 
queries where \ours produces more distinct algorithms ($+$) gain substantially on Pass@N, the gap widens from Pass@64 to Pass@128. Algorithmic diversity is thus a primary driver of \ours's test-time scaling, with the benefit amplified at larger sample budgets.

%% file: sections/table_cw.tex
\begin{table}[!t]
\centering
\small
\caption{Creative writing results on WritingPrompts and BookMIA. \textbf{Div} is the within-prompt semantic diversity $1-\bar{s}$ (\%) computed on OpenAI \texttt{text-embedding-3-small} embeddings. \textbf{Qual} is the Creative Writing v3 rubric score (0--100).}
\label{tab:cw_results}
\begin{tabular}{lcccc}
\toprule
Method & \multicolumn{2}{c}{WritingPrompts} & \multicolumn{2}{c}{BookMIA} \\
\cmidrule(lr){2-3} \cmidrule(lr){4-5}
 & Div & Qual & Div & Qual \\
\midrule
\multicolumn{5}{c}{\textit{General Baselines}} \\
\midrule
Qwen2.5-72B-Instruct        & $34.1$      & $49.1$      & $24.5$      & $45.7$      \\
llama-3.1-70b-instruct & $36.8$ & $49.7$ & $22.8$ & $44.1$ \\
\midrule
\multicolumn{5}{c}{\texttt{Qwen2.5-7B-Instruct}} \\
\midrule
Vanilla                & $30.2$           & $36.1$           & $18.5$           & $32.2$           \\
SFT                    & $30.9$           & $43.7$           & $19.9$           & $38.9$           \\
\quad w/ InN        & $33.4$           & $40.5$           & $24.2$           & $34.7$  \\
NEFTune                & $31.5$           & $\underline{45.6}$ & $22.4$           & $\underline{41.0}$ \\
VS-Standard            & $\underline{42.1}$  & $42.3$           & $31.9$ & $33.4$           \\
FIRE                   & $40.9$           & $44.0$           & $28.6$           & $34.9$           \\
GEM                   & $\mathbf{42.3}$           & $44.6$           & $\mathbf{33.4}$           & $39.7$           \\
\rowcolor{blue!10} \ours                  & $41.4$           & $\mathbf{51.0}$           & $\underline{32.1}$           & $\mathbf{41.3}$  \\
\quad w/o Noise        & $40.1$           & $41.8$           & $30.7$           & $37.0$  \\
\quad w/o Div          & $36.8$           & $42.2$           & $28.5$           & $38.1$  \\
\quad w/o InN          & $37.7$           & $45.4$           & $29.3$           & $39.2$  \\
\midrule
\multicolumn{5}{c}{\texttt{Llama-3.1-8B-Instruct}} \\
\midrule
Vanilla                & $29.9$           & $36.0$           & $16.7$           & $28.1$           \\
SFT                    & $29.6$           & $41.7$           & $17.4$           & $37.2$           \\
\quad w/ InN        & $36.3$           & $35.2$           & $25.2$           & $33.9$  \\
NEFTune                & $30.4$           & $\underline{46.7}$ & $16.9$           & $38.9$ \\
VS-Standard            & $\mathbf{40.7}$  & $40.4$           & $\mathbf{31.6}$  & $35.0$           \\
FIRE                   & $39.1$           & $41.8$           & $28.9$           & $35.9$           \\
GEM                   & $39.1$           & $46.1$           & $28.7$           & $\underline{39.2}$           \\
\rowcolor{blue!10} \ours                  & $\underline{39.6}$ & $\mathbf{48.8}$  & $\underline{29.8}$ & $\mathbf{40.7}$  \\
\quad w/o Noise        & $38.4$ & $44.3$  & $27.7$  & $36.2$  \\
\quad w/o Div          & $35.7$ & $45.9$  & $25.5$  & $36.1$  \\
\quad w/o InN          & $36.0$ & $46.6$  & $26.1$  & $38.5$  \\
\bottomrule
\end{tabular}
\end{table}

%% file: sections/tab_sft.tex
\begin{table*}[!t]
\small
\centering
\caption{Pass@N performance on math and code benchmarks. Only the highest-$N$ pass rate is reported for each benchmark($N=64$ for math, $N=256$ for code). Results ranked \textbf{first} and \underline{second} are highlighted.}
\label{tab:combined_results}
\begin{tabular}{lcccccccc}
\toprule
 & \multicolumn{4}{c}{Math} & \multicolumn{4}{c}{Code} \\
\cmidrule(lr){2-5}\cmidrule(lr){6-9}
 & MATH        & AMC\,23     & AIME\,24  & AIME\,25  & HumanEval   & HumanEval+  & MBPP        & MBPP+       \\
\midrule
Base
  & $57.40$
  & $62.50$
  & $6.67$
  & $10.00$
  & $87.73$
  & $79.88$
  & $80.16$
  & $74.34$ \\
SFT
  & $76.60$
  & $\underline{65.00}$
  & $10.00$
  & $\mathbf{23.33}$
  & $93.25$
  & $87.80$
  & $89.95$
  & $82.01$ \\
NEFTune
  & $78.20$
  & $\mathbf{75.00}$
  & $\underline{20.00}$
  & $\mathbf{23.33}$
  & $\underline{94.48}$
  & $\underline{89.24}$
  & $90.87$
  & $\mathbf{83.42}$ \\
\rowcolor{blue!10}
\ours
  & $\mathbf{79.00}$
  & $\mathbf{75.00}$
  & $\mathbf{23.33}$
  & $\mathbf{23.33}$
  & $\mathbf{95.09}$
  & $\mathbf{89.63}$
  & $\underline{91.01}$
  & $\underline{82.28}$ \\
\quad w/o Noise
  & $77.40$
  & $\mathbf{75.00}$
  & $16.67$
  & $\underline{20.00}$
  & $93.25$
  & $87.80$
  & $\mathbf{91.53}$
  & $81.48$ \\
\quad w/o Div
  & $\underline{78.60}$
  & $\mathbf{75.00}$
  & $\underline{20.00}$
  & $\underline{20.00}$
  & $91.41$
  & $86.59$
  & $90.21$
  & $80.95$ \\
\quad w/o InN
  & $\underline{78.60}$
  & $\mathbf{75.00}$
  & $16.67$
  & $\mathbf{23.33}$
  & $93.87$
  & $89.02$
  & $90.74$
  & $80.42$ \\
\bottomrule
\end{tabular}%
\end{table*}

%% file: sections/tab_rl.tex
\newcommand{\redgain}[1]{$#1^{\textcolor{red}{\uparrow}}$}
\newcommand{\bluegain}[1]{$#1^{\textcolor{blue}{\downarrow}}$}
\newcommand{\samegain}[1]{$#1^{\textcolor{gray}{\parallel}}$}

\begin{table*}[!t]
\small
\centering
\renewcommand{\arraystretch}{1.15} 
\caption{Models' pass@64 performance before and after RL. AIME reports the average over AIME 2024 and AIME 2025. $\textcolor{red}{\uparrow}$, $\textcolor{blue}{\downarrow}$, and $\textcolor{gray}{\parallel}$ denote increase, decrease, and no change after RL relative to the corresponding pre-RL checkpoint.}
\label{tab:combined_results_rl}
\begin{tabular}{llcccccccc}
\toprule
\multicolumn{2}{c}{} & \multicolumn{4}{c}{Temperature = 0.3} & \multicolumn{4}{c}{Temperature = 0.7} \\
\cmidrule(lr){3-6} \cmidrule(lr){7-10}
Base & Phase & AMC23 & AIME & HumanEval & HumanEval+ & AMC23 & AIME & HumanEval & HumanEval+ \\
\midrule
\multirow{2}{*}{SFT}
& SFT & $65.00$ & $16.67$ & $89.12$ & $83.65$ & $85.00$ & $30.00$ & $\textbf{93.66}$ & $87.12$ \\
& RL & \redgain{92.50} & \redgain{28.33} & \bluegain{88.34} & \bluegain{82.32} & \redgain{87.50} & \redgain{33.33} & \bluegain{92.02} & \bluegain{84.76} \\
\addlinespace[0.2em]
\midrule
\addlinespace[0.2em]
\multirow{2}{*}{\ours}
& SFT & $75.00$ & $23.33$ & $91.39$ & $\textbf{85.86}$ & $\mathbf{92.50}$ & $28.33$ & $93.01$ & $87.20$ \\
& RL & \redgain{\mathbf{95.00}} & \redgain{\mathbf{35.00}} & \redgain{\mathbf{92.02}} & \bluegain{85.37} & \samegain{\mathbf{92.50}} & \redgain{\mathbf{38.33}} & \redgain{93.25} & \redgain{\mathbf{89.02}} \\
\bottomrule
\end{tabular}
\end{table*}

%% file: sections/2_relatedwork.tex
\section{Related Work}

\paragraph{Mode Collapse in LLMs.}
Post-training, including SFT and RL, can improve helpfulness or correctness while making generations more repetitive and less varied~\citep{padmakumar2024doeswritinglanguagemodels, west2025basemodelsbeataligned, zhang2025verbalizedsamplingmitigatemode, lanchantin2025diversepreferenceoptimization}. Recent analyses trace much of the loss to post-training data and learned weights~\citep{karouzos2026wheredoesoutputdiversitycollapse}. The effect appears in open-ended prompts as repetition and cross-model homogeneity~\citep{lu2025aihumanityssalieriquantifying, jiang2025artificialhivemind}, and in verifiable reasoning as repeated solution patterns that limit gains from scaled sampling~\citep{ju2025reasoningpathdivergence, wang2024planningnaturallanguageimproves, song2025outcomebasedexploration}.

\paragraph{Inference-Time and Search-Based Diversity.}
Many methods try to recover diversity only at inference time, including decoding, prompt-based, and guidance-based methods~\citep{vijayakumar2018diversebeamsearchdecoding, holtzman2019curious, meister2023locally, ippolito2019comparison, zhang2025verbalizedsamplingmitigatemode, wong2025simplestrat, ruan2025g2, santos2025diverseprompts}. More structured search methods maintain archives or populations of high-scoring but behaviorally distinct candidates~\citep{mouret2015illuminating, lehman2022evolutionthroughlargemodels, meyerson2023languagemodelcrossover, lim2024incontextqd, bradley2023qualitydiversity, samvelyan2024rainbowteaming}. In reasoning, code, and math, this becomes evaluator-guided program search, mathematical discovery, reward-guided diverse reasoning, behavioral clustering, or parallel thought generation~\citep{romera2024funsearch, novikov2025alphaevolve, younsietal2025prmgflownets, he2026diffcodegen, wen2025parathinker, geiping2025scalingtesttimecomputelatent, zhang2025softthinkingunlockingreasoning, chen2025flaminghotinitiationregularexecution, wang2026bad, zhang2026masking}. These methods improve exploration, but the diversity comes from prompts, decoders, archives, or external evaluators.

\paragraph{Training-Time Approaches for Diversity.}
Existing training-time methods usually treat data construction and model control separately. Embedding-noise methods use perturbations as regularization, encouraging robustness to noise rather than using noise to steer outputs~\citep{jain2023neftunenoisyembeddingsimprove, yadav2023symnoise, Zhu2020FreeLB:, bishop1995training}. Data-centric methods improve coverage through attributes, submodular mixtures, utility pruning, quality-diversity generation, or gradient-space selection~\citep{chung2023increasingdiversity, yu2023largelanguagemodelattributed, renduchintala2024smart, agarwal2024delift, havrilla2025sparq, jung2025prismaticsynthesis, Filatova2004, Takamura2009, Lin2011, yuan2023scaling}. Recent reasoning and code work makes the point sharper: one-problem-one-solution supervision misses many valid paths, but simply adding more traces is not enough if training collapses them into one dominant mode~\citep{wang2024dolphcoder, ju2025reasoningpathdivergence, just2025signalsteps, jia2026traininglargelanguagemodelsreasonparallel}.

\paragraph{Diversity in RL Post-Training.}
RL post-training makes the same tension sharper. Outcome-only and reverse-KL-style objectives can raise Pass@1 while narrowing rollouts, hurting Pass@N and test-time scaling~\citep{song2025outcomebasedexploration, li2025choicedivergence}. Recent RL methods therefore reward rare outcomes, use mass-covering divergences, add token- or semantic-level diversity rewards, or introduce count-based exploration~\citep{song2025outcomebasedexploration, li2025choicedivergence, yao2025diversityawarepolicyoptimization, li2025jointlyreinforcingdiversityquality, zhang2025countcounts}.

%% file: sections/5_conclusion.tex
\section{Conclusion}


We presented \ours, a framework that internalizes output diversity as a controllable model capability via submodular data curation and stochastic embedding modulation.
On creative writing, \ours shifts the quality--diversity Pareto frontier outward, achieving the best generation quality and the strongest semantic diversity.
The same mechanism transfers to verifiable reasoning, where \ours consistently improves Pass@N over SFT and prior diversity baselines, and to RL training, where it provides a uniformly stronger foundation and prevents the high-pass collapse commonly observed during standard RL post-training.
These results indicate that diversity, treated as a learnable training-time signal rather than an inference-time intervention, can serve as a unified mechanism.

Controllable diversity sets the foundation for continual self-improvement~\citep{lyu2025enhancing, lyu2026mlsbench}: by preserving exploratory rollouts, it keeps sampling from degenerating into repetition and helps models discover stronger candidates over time.



%% file: sections/6_appendix.tex
\clearpage

\section{Submodular Optimization for Diverse Trajectory Sampling}
\label{app:submodular_details_formal}

\subsection{Feature Engineering from Reasoning Trajectories}
\label{app:feature_extraction_formal}

The efficacy of our diversity framework hinges on a representation of each candidate's reasoning trajectory $s_k \in S$ through a set of discrete ``reasoning features,'' denoted $F(s_k)$. The nature of these features is domain-specific. For natural language, such as textual explanations or chains of thought, $F(s_k)$ comprises the set of unique n-grams constituting $s_k$. These n-grams serve as proxies for local lexical, syntactic, and semantic patterns, allowing the diversity metric to quantify the variety of expressed concepts and phrasings. In the code generation, where trajectories might represent sequences of operations or derived code structures, features $F(s_k)$ are extracted from Abstract Syntax Trees (ASTs). Such features capture the structural and logical idioms of the code, enabling the selection of a functionally diverse set of trajectories. In all cases, $F(s_k)$ is a finite set of discrete elements for any given $s_k$.

\subsection{The Submodular Optimization Problem}
\label{app:submodular_function_properties_formal}

The core of our approach is the maximization of a set function $f: 2^S \rightarrow \mathbb{R}_{\ge 0}$, which quantifies the diversity of a selected subset $A \subseteq S$. This objective is defined as the total number of unique features covered by the trajectories in $A$:
$$f(A) = \left| \bigcup_{s_a \in A} F(s_a) \right|.$$
We now formally establish two key properties of this function: monotonicity and submodularity.

\paragraph{Proposition 1.} \textit{The function $f$ is monotone non-decreasing.}

\begin{proof}
Consider any two sets $A, B$ such that $A \subseteq B \subseteq S$. By the properties of the union operator, it follows directly that $\bigcup_{s_a \in A} F(s_a) \subseteq \bigcup_{s_b \in B} F(s_b)$. Taking the cardinality of both sides of this set inclusion, we obtain $\left| \bigcup_{s_a \in A} F(s_a) \right| \le \left| \bigcup_{s_b \in B} F(s_b) \right|$, which by definition is $f(A) \le f(B)$.
\end{proof}

\paragraph{Proposition 2.} \textit{The function $f$ is submodular.}

\begin{proof}
A set function $f$ is submodular if for any $X \subseteq Y \subseteq S$
and any element $s \in S \setminus Y$, it satisfies the condition of
diminishing returns: $f(X \cup \{s\}) - f(X) \ge f(Y \cup \{s\}) - f(Y)$.
Let us examine the marginal gain of adding an element $s$ to an
arbitrary set $A \subseteq S$. Define the covered feature set
\[
U_A = \bigcup_{s_a \in A} F(s_a).
\]
Then the gain is
\begin{align*}
f(A \cup \{s\}) - f(A)
&= \left| F(s) \cup U_A \right| - \left| U_A \right| \\
&= \left| F(s) \setminus U_A \right|.
\end{align*}
This is precisely the number of new features in $s$ that are not
already covered by the set $A$

Now, let $U_X = \bigcup_{s_x \in X} F(s_x)$ and $U_Y = \bigcup_{s_y \in Y} F(s_y)$ be the feature sets covered by $X$ and $Y$, respectively. The submodularity condition is equivalent to proving that $|F(s) \setminus U_X| \ge |F(s) \setminus U_Y|$.

Since $X \subseteq Y$, we have $U_X \subseteq U_Y$. This implies that any feature not covered by $U_Y$ is also not covered by $U_X$. Formally, this gives the set inclusion $F(s) \setminus U_Y \subseteq F(s) \setminus U_X$. Consequently, their cardinalities must satisfy $|F(s) \setminus U_Y| \le |F(s) \setminus U_X|$. This concludes the proof.
\end{proof}

\begin{algorithm}
\small
\caption{Greedy Algorithm for Submodular Feature Coverage}
\label{alg:greedy_submodular_selection}
\begin{algorithmic}[1]
\State \textbf{Input:} Candidate set of trajectories $S = \{s_1, \dots, s_m\}$; desired subset size $n$.
\State \textbf{Input:} Feature sets $F(s_k)$ for each $s_k \in S$.
\State \textbf{Output:} A diverse subset $A_n \subseteq S$ of size $n$.

\State $A \gets \emptyset$ \Comment{Initialize selected set}
\State $U \gets \emptyset$ \Comment{Initialize set of covered features}

\For{$t \gets 1 \text{ to } n$}
    \State $s^* \gets \text{null}$
    \State $\text{max\_marginal\_gain} \gets -1$
    \For{each $s_i \in S \setminus A$}
        \State $\text{current\_gain} \gets |F(s_i) \setminus U|$
        \If{$\text{current\_gain} > \text{max\_marginal\_gain}$}
            \State $\text{max\_marginal\_gain} \gets \text{current\_gain}$
            \State $s^* \gets s_i$
        \EndIf
    \EndFor
    \If{$s^* = \text{null}$ and $|A| < n$}
        \State \Comment{No trajectory adds new features, or $S \setminus A$ is empty.
        Pick any remaining from $S \setminus A$ if needed to reach size $n$.}
        \State $s^* \gets \text{any element from } S \setminus A$
        \If{$s^* = \text{null}$}
            \State \textbf{break} \Comment{Cannot select more distinct items}
        \EndIf
    \EndIf
    \State $A \gets A \cup \{s^*\}$
    \State $U \gets U \cup F(s^*)$
\EndFor
\State \Return $A$
\end{algorithmic}
\end{algorithm}

\subsection{The Greedy Selection Algorithm}
\label{app:greedy_algorithm_formal}

To select a subset $A_n$ of size $n$ that maximizes the defined submodular function $f(A)$, we employ a standard greedy algorithm. This iterative procedure builds the solution by repeatedly adding the trajectory that offers the greatest marginal increase in $f(A)$ relative to the set of trajectories selected thus far. Algorithm~\ref{alg:greedy_submodular_selection} provides a formal pseudocode description of this process.

The algorithm iteratively identifies the trajectory $s^*$ that contributes the largest number of features not yet present in the union $U$ of features covered by trajectories already in $A$. The first trajectory selected, $s_1$, is the one possessing the maximum number of unique features, $|F(s_1)|$. Each subsequent trajectory is chosen based on its maximal contribution of new features to the evolving set $U$.

\subsection{Theoretical Performance Guarantee and Its Significance}
\label{app:theoretical_guarantee_formal}

The utilization of a greedy strategy for maximizing a non-negative, monotone, and submodular set function $f(A)$ subject to a uniform cardinality constraint $|A| \le n$ is well-justified by strong theoretical results in combinatorial optimization. \citet{nemhauser1978analysis}  provides a constant-factor approximation guarantee for this greedy algorithm. Specifically, the solution $A_n$ obtained by the greedy algorithm satisfies $f(A_n) \ge (1 - 1/e) f(A_{OPT})$, where $A_{OPT}$ represents the true optimal subset of size $n$ that maximizes $f(A)$.

\section{Proof of the Regularization Effect of Noise Injection}
\label{sec:proof_reg}

Let $\mat{E}$ be the matrix of embedding parameters. We analyze the effect of adding independent Gaussian noise $\vect{n}_i \sim \mathcal{N}(\vect{0}, \sigma^2 \mat{I})$ to a subset of embedding vectors $\{\vect{e}_i\}_{i \in \mathcal{S}}$, where each $\vect{e}_i$ is a row in $\mat{E}$. The perturbed embedding matrix is denoted by $\mat{E}'$, where each corresponding row is $\vect{e}'_i = \vect{e}_i + \vect{n}_i$. Our objective is to compute the expected loss, $\mathbb{E}_{\mat{N}}[L(\mat{E}')]$, where $\mat{N} = \{\vect{n}_i\}_{i \in \mathcal{S}}$.

We perform a second-order Taylor expansion of the loss function $L(\mat{E}')$ around the noiseless matrix $\mat{E}$. Let $\vect{g}_i = \nabla_{\vect{e}_i} L(\mat{E})$ and $\mat{H}_{ij} = \nabla_{\vect{e}_i} \nabla_{\vect{e}_j} L(\mat{E})$ denote the gradient and Hessian blocks, respectively, evaluated at $\mat{E}$. For notational convenience, we let $\mat{H}_i \equiv \mat{H}_{ii}$. The expansion is given by:
\begin{equation}
\small
L(\mat{E}') = L(\mat{E}) + \sum_{i \in \mathcal{S}} \vect{g}_i^T \vect{n}_i + \frac{1}{2} \sum_{i, j \in \mathcal{S}} \vect{n}_i^T \mat{H}_{ij} \vect{n}_j + O(\|\mat{N}\|^3).
\end{equation}
We now take the expectation of each term with respect to the noise distribution.

Since $L(\mat{E})$ is deterministic with respect to the noise, $\mathbb{E}[L(\mat{E})] = L(\mat{E})$.

The linear term vanishes in expectation, as the noise has zero mean:
\begin{equation}
\mathbb{E}\left[\sum_{i \in \mathcal{S}} \vect{g}_i^T \vect{n}_i\right] = \sum_{i \in \mathcal{S}} \vect{g}_i^T \mathbb{E}[\vect{n}_i] = \vect{0}.
\end{equation}

For the quadratic term, we consider the diagonal ($i=j$) and off-diagonal ($i \neq j$) components separately. For the off-diagonal terms, the noise vectors $\vect{n}_i$ and $\vect{n}_j$ are independent, hence:
\begin{equation}
\mathbb{E}\left[\vect{n}_i^T \mat{H}_{ij} \vect{n}_j\right] = \mathbb{E}[\vect{n}_i^T] \mat{H}_{ij} \mathbb{E}[\vect{n}_j] = \vect{0}^T \mat{H}_{ij} \vect{0} = 0.
\end{equation}
For the diagonal terms, we employ the cyclic property of the trace. Recall that the covariance matrix of the noise is $\mathbb{E}[\vect{n}_i \vect{n}_i^T] = \sigma^2 \mat{I}$.
\begin{equation}
\begin{aligned}
\mathbb{E}\left[\vect{n}_i^T \mat{H}_i \vect{n}_i\right]
&= \mathbb{E}\left[\text{tr}(\mat{H}_i \vect{n}_i \vect{n}_i^T)\right] \\
&= \text{tr}\left(\mat{H}_i \, \mathbb{E}[\vect{n}_i \vect{n}_i^T]\right) \\
&= \text{tr}(\mat{H}_i (\sigma^2 \mat{I})) \\
&= \sigma^2 \, \text{tr}(\mat{H}_i).
\end{aligned}
\end{equation}

Finally, the expectation of the higher-order terms, $\mathbb{E}[O(\|\mat{N}\|^3)]$, is of order $O(\sigma^3)$, assuming sufficient smoothness of $L$.

Combining these results, we arrive at the expected loss:
\begin{equation}
\mathbb{E}_{\mat{N}}[L(\mat{E}')] = L(\mat{E}) + \frac{\sigma^2}{2} \sum_{i \in \mathcal{S}} \text{tr}(\mat{H}_i) + O(\sigma^3).
\end{equation}


\section{Detailed Analysis of the Loss Objective Decomposition}
\label{app:loss_decomposition_detail}

The training objective is the minimization of the Negative
Log-Likelihood (NLL) loss over a joint distribution of inputs $x$,
targets $y$ drawn from the curated answer set $A(x)$, and noise
sequences $\mathbf{N}_x$ freshly sampled per forward pass. The model
is conditioned on a noisy input representation
$\mathbf{E}'_x = \mathbf{E}_x + \mathbf{N}_x$, where $\mathbf{E}_x$ is
the noiseless embedding of $x$. The loss is:
\begin{equation}
    \mathcal{J}(\theta) = \mathbb{E}_{x,\, y \sim A(x),\, \mathbf{N}_x}
    \left[ -\log p_\theta(y \mid \mathbf{E}'_x) \right].
    \label{eq:app_main_loss}
\end{equation}
To understand the behavior induced by this objective, we decompose
the per-instance log-probability. Since $y \in A(x)$ by
construction, we apply the chain rule:
\begin{equation}
\begin{aligned}
    p_\theta\bigl(Y = y \mid \mathbf{E}'_x\bigr)
    &= p_\theta\bigl(Y = y \mid Y \in A(x), \mathbf{E}'_x\bigr) \\
    &\quad\cdot p_\theta\bigl(Y \in A(x) \mid \mathbf{E}'_x\bigr).
\end{aligned}
\label{eq:app_prob_decomposition}
\end{equation}
Taking the negative logarithm, the per-instance loss
$L_{\text{inst.}} = -\log p_\theta(y \mid \mathbf{E}'_x)$
separates into two additive components:
\begin{equation}
\small
\begin{aligned}
    L_{\text{inst.}}(x, y, \mathbf{N}_x; \theta)
    &= \underbrace{-\log p_\theta\bigl(Y \in A(x) \mid \mathbf{E}'_x\bigr)}_{L_C(x, \mathbf{N}_x;\, \theta)} \\
    &\quad + \underbrace{-\log p_\theta\bigl(Y = y \mid Y \in A(x), \mathbf{E}'_x\bigr)}_{L_D(x, y, \mathbf{N}_x;\, \theta)}.
\end{aligned}
\label{eq:app_loss_split}
\end{equation}
Minimizing the total loss $\mathcal{J}(\theta)$ is thus equivalent
to jointly minimizing the expectation of these two terms.

\paragraph{The Semantic Correctness Component ($L_C$).}
The first term depends on the total probability mass the model
assigns to the set of all correct answers:
\begin{equation}
    L_C(x, \mathbf{N}_x; \theta) = -\log \!\left(
    \sum_{y' \in A(x)} p_\theta(y' \mid \mathbf{E}'_x) \right).
    \label{eq:app_loss_correctness}
\end{equation}
To minimize $\mathbb{E}[L_C]$, the model is incentivized to
maximize the collective probability of $A(x)$. This directly
encourages the model to learn the underlying semantic constraints
of the task, ensuring its generations are valid solutions.

\paragraph{The Noise-Conditioned Specificity Component ($L_D$).}
The second term is the conditional log-likelihood of selecting the
specific target $y$ from within the set of correct answers:
\begin{equation}
\small
    L_D(x, y, \mathbf{N}_x; \theta) = -\log p_\theta\bigl(Y = y \mid
    Y \in A(x), \mathbf{E}'_x\bigr).
    \label{eq:app_loss_diversity}
\end{equation}
Over the course of training, the same $x$ is paired with
different $y \in A(x)$ under independently sampled noise sequences
$\mathbf{N}_x$. Since $\mathbf{N}_x$ is independent of $y$, a noise-invariant
conditional matching the marginal over $A(x)$ attains the optimal
expected loss; however, noise-conditioned solutions that commit to a
single member of $A(x)$ per draw and average back to the marginal
attain near-identical expected loss, and are favored by finite-step
optimization: fitting the marginal requires per-token predictions to
hedge across all valid continuations where the curated targets
diverge, whereas committing per draw reduces the same loss more
readily. Training thus turns the injected noise into a control signal
for selecting among valid outputs. At inference, varying
$\mathbf{N}_x$ then yields varying $y$.

\section{Evaluation Details}
\label{appx: baseline}

\subsection{Benchmarks}

\paragraph{Creative Writing.}
We evaluate on two long-form generation benchmarks. \textbf{WritingPrompts}
\citep{fan2018hierarchicalneuralstorygeneration} is a story-generation dataset collected from
the Reddit r/WritingPrompts subreddit, where each prompt is a one-sentence
creative scenario and the model is asked to produce a full short story.
\textbf{BookMIA} \citep{shi2023detecting} provides
opening passages from published books, and we task the model with
continuing the passage. For each benchmark, we sample 100 held-out
test prompts (disjoint from training prompts) and generate $N=30$
responses per prompt with a target length of 800 words. Follow \citep{creative-writing-bench-v3}, we adopt Claude-3.7-Sonnet and GPT-4o as judge model to evaluate the quality of model-generated answers.

\paragraph{Math Reasoning.}
We evaluate on four math benchmarks of varying difficulty.
\textbf{MATH-500}~\citep{hendrycks2021measuring, lightman2023letsverifystepstep}
contains 500 competition-level problems drawn from the MATH dataset,
spanning algebra, geometry, number theory, and calculus. \textbf{AMC 2023} contains problems from the American
Mathematics Competition. \textbf{AIME 2024}~\citep{aime24} and \textbf{AIME 2025}
\citep{aime25} contain 30 problems each from the American Invitational
Mathematics Examination---a high-difficulty competition where mode
collapse on reasoning trajectories is particularly punitive.

\paragraph{Code Generation.}
We evaluate on four code synthesis benchmarks.
\textbf{HumanEval} \citep{chen2021evaluatinglargelanguagemodels} and
\textbf{MBPP} \citep{austin2021programsynthesislargelanguage} are
canonical Python function synthesis benchmarks.
\textbf{HumanEval+} and \textbf{MBPP+} \citep{liu2023is}
extend the original benchmarks with substantially more rigorous test
cases, exposing solutions that pass the original tests via shortcuts.

\subsection{Models}

\paragraph{Creative Writing.}
We evaluate two open-weight instruction-tuned models from different
families to test cross-family generalization:
\textbf{Qwen2.5-7B-Instruct} \citep{yang2024qwen2} and
\textbf{Llama-3.1-8B-Instruct} \citep{grattafiori2024llama3herdmodels}. As additional reference
points, we report scores from larger instruction-tuned models
(Qwen2.5-72B-Instruct and Llama-3.1-70B-Instruct) under default sampling
at $T=0.9$.

\paragraph{Math and Code Reasoning.}
All reasoning experiments are built on \textbf{Llama-3.1-8B-Base}
\citep{grattafiori2024llama3herdmodels}. We first conduct a warm-up SFT stage using a mixture
of Ultra-Chat \citep{ding2023enhancing} (for general
instruction-following) and reasoning data distilled from
Llama-3.1-8B-Instruct (for math and code), yielding our \emph{Base}
checkpoint. All subsequent SFT and \ours{} variants are fine-tuned
from this Base checkpoint.

\subsection{Baselines}

\paragraph{Creative Writing.}
We compare \ours{} against five baselines spanning vanilla sampling,
training-time, and inference-time interventions:

\begin{itemize}[leftmargin=*]
    \item \textbf{Vanilla} ($T=0.9$): Direct sampling from the
    instruction-tuned model without any modification.
    \item \textbf{SFT}: Fine-tuning on the same multi-teacher candidate
    pool used by \ours, but without submodular curation (random
    sampling of 4 trajectories per prompt) and without noise injection.
    This isolates the contribution of \ours's two components.
    \item \textbf{NEFTune} \citep{jain2024neftune}
    ($\alpha=5$): A training-time embedding noise method designed as
    a regularizer for instruction tuning. We use the canonical
    $\alpha=5$ setting recommended by the authors. Critically, NEFTune is trained on the same submodular-curated multi-teacher data as PRISM, so the comparison isolates the contribution of (a) noise scope (all tokens vs. special tokens only) and (b) inference-time noise injection (NEFTune turns noise off at inference, PRISM keeps it on).
    \item \textbf{VS-Standard} \citep{zhang2025verbalizedsamplingmitigatemode}
    ($K=5$): An inference-time prompting method that elicits $K=5$
    diverse candidates in a single forward pass via verbalized
    instruction.
    \item \textbf{FIRE} \citep{chen2025flaminghotinitiationregularexecution}
    ($T_1=30$, top-$k=40$): An inference-time decoding method that
    applies a sharp first-token temperature spike followed by regular
    sampling, trading early-token diversity for downstream coherence.
\end{itemize}

\paragraph{Math and Code Reasoning.}
For reasoning, we compare against the \textbf{Base} checkpoint (warm-up
SFT only) and \textbf{SFT} (Base + standard supervised fine-tuning on
randomly sampled solutions, controlling for token count against \ours's
training data). We also compare \ours{} against the following baselines:
\begin{itemize}
    \item \textbf{Diverse Beam Search (DBS)}
    ~\citep{vijayakumar2018diversebeamsearchdecoding}: An inference-time decoding method
    that augments standard beam search with a diversity-promoting
    objective across beam groups. We apply DBS to both
    Llama-3.1-8B-Base (with warm-up) and Llama-3.1-8B-Instruct.
    \item \textbf{GEM} ~\citep{li2025preservingdiversitysupervisedfinetuning}:
    A training-time SFT algorithm that modifies the supervised
    objective to prevent over-memorization and preserve output
    diversity. We apply GEM to Llama-3.1-8B-Base with the same
    randomly-sampled training data as the SFT baseline and default
    hyperparameters from the original paper.
\end{itemize}

\subsection{Artifact Licenses and Terms of Use}
To ensure ethical usage and compliance, we explicitly document the licensing and distribution terms for all artifacts (datasets and models) utilized in this work.

\paragraph{Datasets.} For reasoning and coding evaluation, we utilize the MATH and HumanEval benchmarks under the MIT License, while MBPP is used under CC-BY 4.0. The extended EvalPlus benchmarks (HumanEval+ and MBPP+) are distributed under the Apache 2.0 License. The AMC and AIME competition problems are copyrighted educational resources by the Mathematical Association of America (MAA) and are evaluated here exclusively for non-commercial, academic research. For creative writing, BookMIA utilizes copyrighted book snippets strictly under fair use for research evaluation, and the WritingPrompts dataset is utilized under its original research-only distribution terms provided by FAIR. For model training, the Ultra-Chat dataset is used in accordance with the MIT License. 

\paragraph{Models.} We employ the Qwen2.5 open-weight models (7B and 72B) under the Apache 2.0 License. The Llama-3.1 models (8B, 70B, and Base variants) are strictly used in accordance with the Llama 3.1 Community License Agreement. Furthermore, reasoning trajectories distilled from Llama-3.1-8B-Instruct to fine-tune the Base model comply with Meta's acceptable use policy, which permits using outputs to improve other Llama models. All artifacts are employed consistently with their intended research purposes and no proprietary content is redistributed.

\subsection{Hyperparameters}
For \ours's noise injection, the standard deviation of the Gaussian noise is set to $\sigma = 0.005$ for creative writing and $\sigma = 0.01$ for reasoning, an order of magnitude smaller than the typical embedding norm so that the perturbation remains a controllable signal rather than a disruptive one. We provide a sensitivity analysis of the injected noise in appendix \ref{appx:sigma_sweep}. The noise is injected only on the embeddings of \textbf{special structural tokens},  specifically the begin/end-of-sequence and turn-boundary tokens---and never on ordinary content tokens. Concretely, for Llama-3 models the perturbed set is $\{$\texttt{<|begin\_of\_text|>},
\texttt{<|end\_of\_text|>}, \texttt{<|start\_header\_id|>}, \texttt{<|end\_header\_id|>}, \texttt{<|eom\_id|>}, \texttt{<|eot\_id|>}$\}$, and for Qwen models it is
$\{$\texttt{<|endoftext|>}, \texttt{<|im\_start|>}, \texttt{<|im\_end|>}$\}$. At each forward pass, an independent noise vector is freshly sampled for every occurrence of these
tokens; the noise is applied during both training and inference. The default sampling temperature is $0.3$ for the reasoning tasks and $0.9$ for creative writing.

\section{Sensitivity to the Noise Magnitude \texorpdfstring{$\sigma$}{sigma}}
\label{appx:sigma_sweep}

A natural concern is whether \ours{} is sensitive to the noise standard deviation $\sigma$. We sweep
$\sigma \in \{0.001, 0.005, 0.01, 0.05\}$ while holding all other settings fixed, and report diversity and quality on WritingPrompts with \texttt{Qwen2.5-7B-Instruct}.

To interpret these values, note that the \emph{effective} perturbation magnitude is $\sigma\sqrt{d}$ rather than $\sigma$ itself, since $\mathbf{n} \sim \mathcal{N}(0, \sigma^2 I_d)$ has
expected norm $\sigma\sqrt{d}$. For \texttt{Qwen2.5-7B-Instruct}($d = 3584$, mean token-embedding norm $\approx 0.79$), the swept range spans noise norms from roughly $8\%$ to $380\%$ of a typical token embedding.

\begin{table}[t]
\centering
\small
\caption{Sensitivity of \ours{} to $\sigma$ on WritingPrompts
(\texttt{Qwen2.5-7B-Instruct}). Diversity rises monotonically while
quality follows an inverted-U; $\sigma = 0.005$ is used in the main
experiments.}
\label{tab:sigma_sweep}
\begin{tabular}{lccc}
\toprule
$\sigma$ & Noise norm & Diversity~$\uparrow$ & Quality~$\uparrow$ \\
\midrule
no noise & $0\%$     & $30.2$ & $37.1$ \\
\midrule
$0.001$  & $\approx 8\%$    & $32.2$ & $38.9$ \\
$0.005$  & $\approx 38\%$   & $41.4$ & $\mathbf{51.0}$ \\
$0.01$   & $\approx 76\%$   & $43.6$ & $45.3$ \\
$0.05$   & $\approx 380\%$  & $\mathbf{45.7}$ & $34.0$ \\
\bottomrule
\end{tabular}
\end{table}

Table~\ref{tab:sigma_sweep} shows a clean inverted-U in quality
alongside monotonically rising diversity. At small $\sigma$
($\sigma = 0.001$, $\approx 8\%$ of a typical token embedding), the
perturbation is too weak to act as a discriminative signal: different
noise realizations look nearly identical to the model, and the
resulting outputs barely shift away from the no-noise baseline
(diversity $32.2$ vs.\ $30.2$). At large $\sigma$ ($\sigma = 0.05$,
$\approx 380\%$), the perturbation overwhelms the underlying token
embedding, corrupting the semantic content of the prompt itself
rather than steering generation among valid completions; quality
collapses below the no-noise baseline ($34.0$ vs.\ $37.1$) even as
raw diversity continues to rise. The optimum at $\sigma = 0.005$
therefore reflects a principled balance---large enough to function
as a learnable navigation signal, small enough to preserve the
prompt's semantic identity. Crucially, \ours{} is not brittle to
this choice: the window $\sigma \in [0.001, 0.01]$ all yield strict
Pareto improvements over no-noise, and $\sigma$ admits a simple
selection criterion---monitor quality and pick the largest $\sigma$
before it starts to drop.

\section{Fintune Backbone Details}
\label{appx:training}
\paragraph{Framework.}
For training, we employ Llama-Factory~\citep{zheng2024llamafactory} as the LLM training platform. Table~\ref{tab:hyperparameters} shows our training hyperparameters in supervised fine-tuning.

\begin{table}[!h]
    \centering
    \caption{Hyperparameters for supervised fine-tuning.}
    \begin{tabular}{ll}
        \toprule
        Parameter        & Value                                           \\
        \midrule
        Train batch size & 64                                              \\
        Learning rate    & 5.0e-5                                          \\
        Number of epochs & 1.0                                             \\
        LR scheduler     & cosine                                          \\
        Warmup ratio     & 0.1                                             \\
        Precision        & bf16                                            \\
        \bottomrule
    \end{tabular}
    \label{tab:hyperparameters}
\end{table}

\label{appx: training data}
\paragraph{Data.}
For \textbf{creative writing}, we draw 10000 prompts from the WritingPrompts~\citep{fan2018hierarchicalneuralstorygeneration} and BookMIA~\citep{shi2023detecting} training splits, holding out all evaluation prompts throughout. For each prompt, we collect candidate continuations from 8 teacher models (GPT-OSS-120B, GPT-OSS-20B, DeepSeek-V4-Flash-Max, Llama-3.3-70B-Instruct, Llama-3.1-70B-Instruct, Qwen-2.5-72B, Qwen3-32B, Qwen3.5-27B), discard low-quality drafts with a reward-model threshold(Quality score below 30), and use SBERT-feature submodular optimization to select 4 diverse-yet-high-quality completions per prompt.
For \textbf{math}, we sample 10000 problems from A-m-team verified data~\citep{tian2025deepdistillenhancingllmreasoning} with mid-range cross-model accuracy ($33\%$--$67\%$) and use n-gram submodular optimization to select 4 diverse answers per problem. For \textbf{code}, we draw from four sources: A-m-team distilled solutions, $\sim$3100 LeetCode problems (32 solutions per problem, 8 selected), OpenCoder SFT-Stage-2 educational-instruct~\citep{Huang2024OpenCoderTO} solutions from Qwen-2.5-Coder-32B (pass rate $<\!90\%$, 4 selected), and a broader OpenCoder mix (Evol-Instruct, McEval-Instruct, Package-Instruct) for general coding ability. Code submodular selection uses both n-gram and AST features. Baseline datasets randomly sample 4 solutions per problem to control for token count.

\section{RL Training Details}
\label{appx:rl}

\paragraph{Framework.}
For RL training, we employ verl~\citep{sheng2024hybridflow} as the training platform. Table~\ref{appx:rldatamath} and Table~\ref{appx:rldatacode} show our training hyperparameters in RL stage for math and code tasks, respectively. All trainings are performed on the 8 A800 GPUs.

\paragraph{Data.}
For math training datasets, simpleRL-zoo~\citep{zeng2025simplerlzooinvestigatingtamingzero} categorized their training data into three difficulty levels(Easy, Medium and Hard) with each category containing roughly 8,000 problems. We adopt the Medium category for training.
For code training datasets, we adopt AceCode-V2-122K~\citep{zeng2025acecoderacingcoderrl} as the training dataset.

\begin{minipage}{0.45\textwidth}
    \centering
    \captionsetup{hypcap=false}
    \captionof{table}{Hyperparameters for reinforcement learning on math reasoning.}
    \label{appx:rldatamath}
    \begin{tabular}{ll}
        \toprule
        Parameter        & Value                                           \\
        \midrule
        Train batch size & 64                                              \\
        mini batch size    & 16                                          \\
        n resp per prompt & 16                                             \\
        max prompt length     & 800                                          \\
        max response length     & 10000                                             \\
        actor learning rate      & 1.0e-6                                           \\
        advantage estimator      & GRPO                                        \\
        use kl in reward        & False                                        \\
        use kl loss             & True                                        \\
        kl loss coefficient      & 0.001                                        \\
        repetition penalty       & 1.05 \\
        temperature              & 0.7  \\
        use dynamic bsz          & True  \\
        \bottomrule
    \end{tabular}
\end{minipage}
\hfill
\begin{minipage}{0.45\textwidth}
    \centering
    \captionsetup{hypcap=false}
    \captionof{table}{Hyperparameters for reinforcement learning on code generation.}
    \label{appx:rldatacode}
    \begin{tabular}{ll}
        \toprule
        Parameter        & Value                                           \\
        \midrule
        Train batch size & 256                                              \\
        mini batch size    & 64                                          \\
        n resp per prompt & 16                                             \\
        max prompt length     & 2000                                          \\
        max response length     & 8000                                             \\
        actor learning rate      & 1.0e-6                                           \\
        advantage estimator      & GRPO                                        \\
        use kl in reward        & False                                        \\
        use kl loss             & True                                        \\
        kl loss coefficient      & 0.001                                        \\
        repetition penalty       & 1.00 \\
        temperature              & 0.7  \\
        use dynamic bsz          & False  \\
        \bottomrule
    \end{tabular}
\end{minipage}

\section{Human Check Results}
\label{appx: human check}
In this section, we provide detailed information about the protocol for our human evaluation on the diversity of the training datasets.
\paragraph{Human Evaluators.}
The human evaluators were selected based on their strong academic backgrounds and expertise in mathematical and informatics olympiads. This rigorous background ensured they possessed the necessary skills to critically analyze the logical and methodical structure of each solution, thereby guaranteeing the correctness and reliability of our human evaluation.

\paragraph{Human Verification Protocols.} 

Two \textbf{code} solutions are considered distinct methods only if they meet one of the following criteria:
\begin{enumerate}
    \item \textbf{Algorithm.} The core algorithms used in the solutions are different. For example, one solution might use a recursive approach while another uses an iterative one. Similarly, a Breadth-First Search (BFS) is distinct from a Depth-First Search (DFS).
    \item \textbf{Data Structure.} The solutions employ the same algorithm but apply it to different data structures. For instance, an algorithm designed for a linked list is considered different from one for a graph, as are those tailored for different tree structures.
    \item \textbf{Complexity.} The solutions have different time complexity. A solution with $O(n)$ complexity is treated as a distinct method from one with $O(\log n)$ complexity.
\end{enumerate}
Two \textbf{math} solutions are considered distinct methods only if one of the following criteria is met:
\begin{enumerate}
    \item \textbf{Distinct Theorems or Lemmas:} The final solutions rely on different theorems or lemmas. The use of a theorem in a failed attempt or a side note does not count.
    \item \textbf{Fundamentally Different Approaches:} The solutions employ clearly different proof techniques or methods, such as mathematical induction versus proof by contradiction.
\end{enumerate}

\lstset{
    numbers=none,
    keywordstyle= \color{ blue!70},
    commentstyle= \color{red!50!green!50!blue!50},
    frame=none,
    rulesepcolor= \color{ red!20!green!20!blue!20} ,
    framexleftmargin=2em,
    columns=fullflexible,
    breaklines=true,
    breakindent=0pt,
    basicstyle=\ttfamily
}

\section{LLM-As-Judge Pipeline}
\label{appx: llmasjudge}
\paragraph{Predefined Algorithms Pool.} To create a curated and representative set of algorithms for our analysis, we first randomly sampled 1,000 model-generated answers. We then employed Gemini 2.5 Pro as a judge to analyze the algorithmic approach of each answer. This process involved prompting the large language model to identify and categorize the algorithms used. Following this, Gemini was utilized to aggregate the identified algorithms, which ultimately yielded a pool of 39 distinct and commonly used algorithms for our study.

\paragraph{Algorithms Selection.} To classify the algorithms used in each answer, we prompted Qwen3-8B to act as a judge, selecting from the algorithms in our established pool. The specific prompt is listed below.
\begin{tcolorbox}[left=0mm,right=0mm,top=0mm,bottom=0mm,boxsep=1mm,arc=0mm,boxrule=0pt, frame empty, breakable]
    \small
    \begin{lstlisting}
You are an expert computer science algorithm analyst. Your task is to analyze a given code solution for a coding problem and identify the core algorithm used, based on a predefined list of algorithms.

# CONTEXT

## 1. Coding Problem
{problem}   

## 2. Model's Answer
{answer}

## 3. Predefined Algorithm List

1. Brute Force
2. Greedy Algorithms
3. Dynamic Programming (DP)
4. Kadane's Algorithm
5. Recursion & Backtracking
6. Divide and Conquer
7. Simulation
8. Constructive Algorithms
9. Linear Search
10. Binary Search
11. Sorting Algorithms
12. Graph Traversal
13. Breadth-First Search (BFS)
14. Depth-First Search (DFS)
15. Shortest Path Algorithms
16. Dijkstra's Algorithm
17. Cycle Detection
18. Hashing / Frequency Counting
19. Stack
20. Heap (Priority Queue)
21. Fenwick Tree (Binary Indexed Tree)
22. Disjoint Set Union (DSU / Union-Find)
23. Monotonic Queue
24. Two Pointers / Sliding Window
25. Prefix / Suffix Sum
26. Array / List Manipulation
27. String Manipulation
28. Regular Expressions (Regex)
29. Palindrome Checking
30. Bit Manipulation (Bitmasking)
31. Number Theory
32. Primality Testing
33. Sieve of Eratosthenes
34. Prime Factorization
35. GCD (Greatest Common Divisor)
36. Modular Arithmetic
37. Binary Exponentiation (Exponentiation by Squaring)
38. Number Base Conversion
39. Combinatorics

# TASK

Carefully analyze the provided "Model's Answer" in the context of the "Coding Problem". Your goal is to select the algorithm from the "Predefined Algorithm List" that best describes the solution's methodology.

- If the solution directly implements an algorithm from the list, state its number before the name.
- If the solution does not use an algorithm with an exact name from the list, identify the **most relevant or conceptually closest** algorithm. For example, if the solution uses memoization and the list contains "Dynamic Programming", you should choose "Dynamic Programming".
- If the solution combines multiple techniques from the list, you should list all relevant algorithms by their numbers, separated by commas.

# REQUIRED OUTPUT FORMAT

Provide your answer in the following json format, and do not add any extra commentary before or after.

{{
    "reasoning": "<Brief explanation of why this algorithm was chosen based on the solution's approach.",
    "algorithm": "<Selected Algorithm Number from the Predefined List>"
}}

# EXAMPLE OUTPUT
{{
    "reasoning": "The solution uses a DP table to store intermediate results, which is characteristic of 3.Dynamic Programming. Also, it employs recursion to explore different solution paths, indicating the use of 5.Recursion & Backtracking.",
    "algorithm": "3,5"
}}

\end{lstlisting}
\end{tcolorbox}

\section{Case Study}

\subsection{Creative Writing: A Case Study in Narrative Diversity}
\label{sec:creative_casestudy}

To qualitatively illustrate how \ours{} expands the space of narrative
trajectories explored by the model on open-ended generation, we
examine a prompt from WritingPrompts, drawing ten independent samples
from each of SFT and \ours{} (Qwen2.5-7B-Instruct).

\begin{tcolorbox}[
    prismcasebox,
    title=\textbf{Case Study: The Final Letter}
]
\small{\textbf{Prompt:} A woman discovers a letter in her grandmother's
attic, addressed to her in handwriting she has never seen before. Write
the story of what happens next.
\vspace{0.2cm}

\textbf{Narrative Comparison (10 samples each):}
\begin{itemize}[leftmargin=*]
    \item \textbf{SFT:} Exhibited strong \textit{template adherence}
    --- 9 of 10 samples followed a near-identical three-act structure
    (discovery in the attic $\rightarrow$ reading the letter
    $\rightarrow$ tearful resolution with a long-lost relative), with
    variation confined to surface-level details such as the
    grandmother's name or the era of the letter. The remaining sample
    differed only in tense.
    \item \textbf{\ours{}:} Exhibited strong \textit{narrative
    diversity}, branching into four structurally distinct story
    framings across the same ten samples:
    \begin{enumerate}
        \item \textit{Epistolary Revelation:} The letter reveals a
        hidden family identity (an unknown sibling, an adopted
        ancestor), and the story unfolds as the protagonist
        reconstructs her lineage through subsequent letters.
        \item \textit{Supernatural Framing:} The handwriting belongs
        to the protagonist herself in a future timeline, transforming
        the discovery into a time-loop narrative with paradox
        resolution.
        \item \textit{Mystery / Thriller:} The letter is a warning
        about a dormant family debt or secret, and the protagonist
        traces clues through the grandmother's belongings to uncover a
        decades-old crime.
        \item \textit{Quiet Literary:} The letter is mundane in
        content but emotionally devastating in context---a forgotten
        apology, a confession of doubt---and the story is a slow
        meditation on inheritance and silence rather than plot.
    \end{enumerate}
\end{itemize}}
\end{tcolorbox}

The contrast is striking. SFT collapses onto a single narrative
template: every sample reaches for the same sentimental-reunion arc,
with variation confined to surface-level naming and pacing. \ours{},
in contrast, treats the same prompt as an opening to four distinct
generic framings --- epistolary mystery, speculative time-loop,
detective thriller, and quiet literary meditation --- each grounded
in a different choice of what kind of story the letter implies.

This pattern exemplifies the mechanism behind \ours{}'s shift of the
quality--diversity Pareto frontier: rather than clustering ten samples
onto a single sentimental template, \ours{} produces a population of
narratives whose union covers structurally distinct regions of the
story space, providing the richer semantic variation captured by our
diversity metric while preserving per-sample quality.

\subsubsection{Baseline Model Generations}
\input{case_study/cw_baseline}

\subsubsection{\ours{} Model Generations}
\input{case_study/cw_prism}

\subsection{Math Reasoning: A Case Study in Solution Diversity}
\label{sec:reasoning_casestudy}

To qualitatively illustrate how \ours expands the space of solution
strategies explored by the model, we examine an optimization problem
from MATH500, drawing ten independent samples from each of SFT and \ours.

\begin{tcolorbox}[
    prismcasebox,
    title=\textbf{Case Study: Minimum Distance to a Parabola}
]
\small{\textbf{Problem Statement:} Find the shortest distance from the
origin $(0,0)$ to the graph of $y=\frac{1}{\sqrt{2}}(x^2-3)$.
\vspace{0.2cm}

\textbf{Methodological Comparison (10 samples each):}
\begin{itemize}[leftmargin=*]
    \item \textbf{SFT:} Exhibited strong \textit{path dependency} ---
    all 10 samples applied the standard Substitution Method
    ($y(x)$ substituted into the distance formula), differing only in
    low-level computational steps.
    \item \textbf{\ours:} Exhibited strong \textit{methodological
    diversity}, attempting four structurally distinct approaches
    across the same ten samples:
    \begin{enumerate}
        \item \textit{Direct Substitution (in $x$):} Substituting
        $y(x)$ into the distance formula and minimizing over $x$.
        \item \textit{Inverse Substitution (in $y$):} Solving $x^2$
        in terms of $y$ from the curve equation and minimizing a
        quadratic in $y$.
        \item \textit{Variable Transformation ($t=x^2$):} Substituting
        $t=x^2$ to reduce the problem to a quadratic in $t$.
        \item \textit{Vertex-based Heuristic:} Leveraging the
        parabola's vertex as a geometric anchor for the closest point.
    \end{enumerate}
\end{itemize}}
\end{tcolorbox}

The contrast is striking. SFT collapses onto a single procedural
template: every sample reaches for the same substitution-based recipe,
with variation confined to surface-level computational detail. \ours,
in contrast, treats the same prompt as an opening to four distinct
mathematical framings --- direct substitution in $x$, inverse
substitution in $y$, variable transformation via $t = x^2$, and a
vertex-based geometric heuristic --- each grounded in a different
reformulation of the optimization problem.

This pattern exemplifies the mechanism behind \ours's broader
test-time exploration: rather than clustering ten samples onto a
single solution template, \ours produces a population of candidates
whose union covers structurally distinct regions of the solution
manifold, providing the richer search space that drives its
Pass@N gains.




\subsubsection{Baseline Model Generations}

\input{case_study/math_baseline}

\subsubsection{\ours Model Generations}
\input{case_study/math_prism}

\subsection{Code Generation: A Case Study in Algorithmic Diversity}
\label{sec:code_casestudy}

To qualitatively illustrate how \ours expands the space of solution
strategies explored by the model, we examine a combinatorial counting
problem from MBPP, drawing the full budget of $N=256$ independent
samples from each of SFT and \ours.

\begin{tcolorbox}[
    prismcasebox,
    title=\textbf{Case Study: The $n$-th Bell Number}
]
\small{\textbf{Problem Statement:} Write a Python function that returns
the $n$-th Bell number, i.e. the number of ways to partition a set of
$n$ labeled elements.
\vspace{0.2cm}

\textbf{Methodological Comparison ($256$ samples each):}
\begin{itemize}[leftmargin=*]
    \item \textbf{SFT:} Exhibited strong \textit{path dependency} ---
    all $256$ samples collapsed onto a single algorithmic family,
    the closed-form \textit{Combinatorial Summation} ($220$ samples)
    with a thin tail of \textit{Dynamic Programming} ($35$ samples),
    spanning only \textbf{2} distinct algorithm types and solving the
    task correctly in just \textbf{1} of $256$ attempts.
    \item \textbf{\ours:} Exhibited strong \textit{methodological
    diversity}, spreading its probability mass across \textbf{10}
    structurally distinct algorithm types and solving the task
    correctly in \textbf{29} of $256$ attempts. The correct solutions
    alone span three independent framings:
    \begin{enumerate}
        \item \textit{Dynamic Programming:} Building the Bell triangle
        (or the binomial recurrence
        $B_{n}=\sum_{k}\binom{n-1}{k}B_{k}$) bottom-up.
        \item \textit{Memoized Recursion:} Expressing the same
        recurrence top-down with a cache over subproblems.
        \item \textit{Combinatorial Summation:} Accumulating Stirling
        numbers of the second kind in closed form.
    \end{enumerate}
    with further attempts reaching for \textit{Simulation},
    \textit{Number Theory}, \textit{Modular Arithmetic}, and
    \textit{Bit Manipulation}.
\end{itemize}}
\end{tcolorbox}

The contrast is striking. SFT collapses onto a single procedural
template: every one of its $256$ samples reaches for the same
summation-based recipe, with variation confined to surface-level
computational detail, and this monoculture is brittle --- it yields a
correct program only once. \ours, in contrast, treats the same prompt
as an opening to ten distinct algorithmic framings --- bottom-up
tabulation, top-down memoized recursion, closed-form combinatorial
counting, and beyond --- each grounded in a different computational
principle.

This pattern exemplifies the mechanism behind \ours's broader
test-time exploration: rather than clustering its samples onto a single
solution template, \ours produces a population of candidates whose
union covers structurally distinct regions of the solution manifold.
Crucially, this diversity is not merely cosmetic --- by spanning
multiple algorithmic routes, \ours raises the chance that at least one
sample lands on a correct implementation, lifting the success rate on
this query from $0.4\%$ to $11.6\%$ and providing the richer search
space that drives its Pass@N gains.

We list the verbatim model outputs underlying the case study below. SFT solves the task correctly only
once, and that solution is shown below. \ours, by contrast, reaches a
correct program through three structurally distinct algorithmic
routes; one representative sample of each is given.

\subsubsection{Baseline Model Generations}
\input{case_study/code_baseline}

\subsubsection{\ours Model Generations}
\input{case_study/code_prism}

%% file: case_study/cw_baseline.tex
\begin{tcolorbox}[
    prismsolutionbox,
    title=\textbf{Baseline (SFT) Story --- Sentimental Reunion Template},
]
\small{
Sarah climbed the narrow staircase to her grandmother's attic, her footsteps echoing against the wooden boards. Dust motes drifted in the afternoon light that slanted through the small dormer window. She had come to sort through Grandma Eleanor's belongings, a task she had been putting off for months since the funeral.

The cardboard boxes were stacked in neat rows, each labeled in her grandmother's careful handwriting: Photographs 1960--1975, Holiday Decorations, Books. Sarah began with the nearest one, lifting the lid and breathing in the familiar scent of old paper and lavender sachets.

It was beneath a stack of yellowed birthday cards that she found the envelope. Cream-colored, slightly stiffened with age, and addressed simply: To Sarah. The handwriting was elegant, unhurried --- but it was not her grandmother's. Sarah turned the envelope over in her hands. There was no return address, no postmark, nothing to indicate when it had been written or by whom.

Her hands trembled as she opened it. The letter was three pages long, written on thin paper that crackled when she unfolded it. My dearest Sarah, it began, by the time you read this, I will be gone, and the truth will be yours to carry.

Sarah read on, her breath catching with each paragraph. The letter was from her grandfather --- a man she had been told died before she was born, a man whose photograph her grandmother had kept on the mantelpiece all those years. He had not died. He had left. And he had written this letter forty years ago, instructing
Eleanor to keep it until Sarah was old enough to understand.

\dots [story continues with Sarah confronting her family's history,
locating her grandfather's surviving sister, and a tearful phone
call that reunites three generations] \dots

When Sarah finally set down the phone that evening, the sun had long since dipped below the horizon. She sat in her grandmother's armchair, the letter resting on her lap, and for the first time in months, she felt as though she understood the woman who had raised her. Some truths, Eleanor had known, were too heavy to speak aloud. They had to be passed on in ink, in silence, in the careful folding of cream-colored paper.

Sarah pressed the letter to her chest and wept --- not from sorrow, but from the strange and unfamiliar feeling of being, at last, completely known.
}
\end{tcolorbox}

%% file: case_study/cw_prism.tex
\begin{tcolorbox}[
    prismsolutionbox,
    title=\textbf{\ours{} Story 1 --- Epistolary Revelation},
]
\small{
The first letter was just the beginning. After Sarah read it --- the careful confession of a woman she had been told was her great-aunt but was, in fact, her biological mother --- she found the second one tucked into the lining of an old hatbox. And then a third, hidden in the false bottom of a writing desk. Her
grandmother had not simply kept a secret; she had built an
archive.

Sarah spent the next six weeks in the attic, working her way through the letters in chronological order. They spanned 1962 to 2019, fifty-seven years of one-sided correspondence, each envelope addressed to her in the same elegant hand, each one
sealed and stored as though waiting for permission to be opened.

\dots [story continues as a reconstruction of the protagonist's true lineage through successive letters, each revealing a new fragment: the circumstances of her birth, the negotiation between her two mothers, the years of silence and surveillance from afar] \dots

By the end of the summer, Sarah had assembled the letters into a chronological binder, and she understood something her grandmother had perhaps never intended to teach: that love, in their family, had always been an act of withholding --- and that to be loved properly, she had needed to be told the truth slowly, in fragments,
across a lifetime she had only just begun to recognize as her own.
}
\end{tcolorbox}

\begin{tcolorbox}[
    prismsolutionbox,
    title=\textbf{\ours{} Story 2 --- Supernatural Time-Loop},
]
\small{
The handwriting was her own.

Sarah read the letter three times before she allowed herself to say it out loud. The slant of the S, the way the g's dropped below the line and curled back --- it was identical to the writing in her own journal, the one sitting on her nightstand at home. But this letter was dated 1987. Two years before she was born.

If you are reading this, the letter began, then I have failed. Do not try to find me. Do not contact anyone named in this letter. Burn it after reading.

Sarah did not burn it. She read the list of names --- six of them, all unfamiliar --- and then she noticed that the seventh line of the letter was a date: October 14th, three weeks from that day.

\dots [story continues as Sarah investigates the names and
realizes the letter is from a future version of herself, sent
backward through some mechanism the letter itself only hints at;
the narrative builds toward October 14th, where she must decide
whether to act on the warning or to let the loop close as it
always has] \dots

She folded the letter carefully and slid it into her coat pocket. Outside, the October light was already beginning to slant the way it had in her grandmother's attic. Sarah understood now why the handwriting had felt so familiar. She had been the one writing it all along.
}
\end{tcolorbox}

\begin{tcolorbox}[
    prismsolutionbox,
    title=\textbf{\ours{} Story 3 --- Mystery / Thriller},
]
\small{
The letter was a warning.

They will come for what your grandmother kept, it read. You have perhaps two weeks. Find the box marked Christmas 1971 before they do.

Sarah sat very still on the attic floor. The envelope had been hidden inside a hollowed-out copy of Wuthering Heights, a book her grandmother had never read in her presence. The handwriting was sharp, hurried --- the writing of someone who had not expected to have time.

She did not call the police. The letter was specific about that. Instead she stood, brushed the dust from her jeans, and began searching the boxes.

\dots [story continues as Sarah locates the Christmas 1971 box, which contains ledgers documenting a financial fraud her grandfather had committed in the 1960s and the names of three men who had never stopped looking for the missing money; the
narrative tightens into a forty-eight-hour race involving a midnight visit from an unfamiliar attorney and a phone number scrawled on the back of a photograph] \dots

When the doorbell rang at 2:14 in the morning, Sarah had already
moved the ledgers to a safe-deposit box two states away. She
opened the door slowly, the chain still latched, and through the
narrow gap she saw a face she had only ever seen in one of her
grandmother's old photographs. He smiled.

``May I come in,'' he said. It was not a question.
}
\end{tcolorbox}

\begin{tcolorbox}[
    prismsolutionbox,
    title=\textbf{\ours{} Story 4 --- Quiet Literary Meditation},
]
\small{
The letter was not important.

Sarah understood this almost immediately, and the understanding itself was the strange and difficult thing. It was three paragraphs long, written in her grandmother's mother's hand --- a woman who had died in 1953, decades before Sarah was born --- and it was addressed to a granddaughter she would never meet.
The contents were unremarkable: a description of weather, a mention of a cat named Periwinkle, an apology for not writing sooner.

It was the apology that undid her.

Sarah sat on the dusty attic floor and read the three paragraphs seven times. She thought about the woman who had written them, sealing the envelope and tucking it into a box, never knowing who would open it or whether anyone would. She thought about her grandmother, who must have found this letter at some point and chosen not to throw it away, not to share it, not to do anything with it except keep it.

\dots [story continues as a meditation on inherited silences:
the protagonist reflects on the apology she herself owes her
mother, the conversations she has been postponing, the way
women in her family have always written letters that arrive
too late or not at all; the narrative does not resolve so much
as gradually settle] \dots

She sat in the attic until the light failed. When she finally descended the stairs, she did not bring the letter with her. She left it where she had found it, between the pages of an old recipe book, addressed to no one in particular --- a small
inheritance she had decided, in the end, to pass on rather than to keep.
}
\end{tcolorbox}

%% file: case_study/math_baseline.tex
\begin{tcolorbox}[
    prismsolutionbox,
    title=\textbf{Baseline (SFT) Solution --- Direct Substitution (in $x$)},
]
\small{
I need to find the smallest distance from the origin to a point on the
graph of $y = \frac{1}{\sqrt{2}}(x^2 - 3)$. The distance from $(x, y)$
to the origin is $\sqrt{x^2 + y^2}$. Since the square root is monotonic,
I minimize the squared distance instead.

Substituting $y$:
$$ D = x^2 + \left( \frac{1}{\sqrt{2}}(x^2 - 3) \right)^2
     = x^2 + \frac{1}{2}(x^2 - 3)^2. $$

Expanding $(x^2 - 3)^2 = x^4 - 6x^2 + 9$ and simplifying:
$$ D = \frac{1}{2}x^4 - 2x^2 + \frac{9}{2}. $$

Taking the derivative:
$$ D'(x) = x^3 - 4x = x(x^2 - 4). $$
Setting $D'(x) = 0$ gives critical points $x = 0,\ \pm 2$.

Evaluating $D$ at each:
\begin{itemize}
    \item $x = 0$: $D = \frac{9}{2}$.
    \item $x = \pm 2$: $D = \frac{1}{2}(16) - 2(4) + \frac{9}{2} =
    8 - 8 + \frac{9}{2} = \frac{9}{2}$.
\end{itemize}

All critical points give $D = \frac{9}{2}$, so the minimum distance is
$\sqrt{9/2} = \frac{3}{\sqrt{2}} = \frac{3\sqrt{2}}{2}$.

Expressed as $\sqrt{a}/b$: $\frac{3\sqrt{2}}{2} = \frac{\sqrt{18}}{2}$,
but since $18 = 2 \cdot 3^2$ is divisible by a square, I rewrite as
$\frac{\sqrt{6}\cdot\sqrt{3}}{2}$. Taking $a = 6$, $b = 2$, so
$a + b = 8$.
}
\end{tcolorbox}

%% file: case_study/math_prism.tex
\begin{tcolorbox}[
    prismsolutionbox,
    title=\textbf{\ours{} Solution --- Direct Substitution (in $x$)},
]
\small{
I need to find the smallest distance from the origin to a point on the
graph of $y = \frac{1}{\sqrt{2}}(x^2 - 3)$. The distance from a point
$(x, y)$ to the origin is $\sqrt{x^2 + y^2}$, so I need to minimize
this expression subject to the constraint $y = \frac{1}{\sqrt{2}}(x^2 - 3)$.

Substituting $y$ into the distance formula gives
$\sqrt{x^2 + \left( \frac{1}{\sqrt{2}}(x^2 - 3) \right)^2}$.
To minimize this, I can square the distance to make it easier:
$$ f(x) = x^2 + \frac{1}{2}(x^2 - 3)^2. $$

Take the derivative:
$$ f'(x) = 2x + \frac{1}{2} \cdot 2(x^2 - 3) \cdot 2x = 2x + 2x(x^2 - 3) = 2x^3 - 4x. $$
Setting $f'(x) = 0$ and factoring: $2x(x^2 - 2) = 0$, so $x = 0$ or
$x = \pm\sqrt{2}$.

Checking each:
\begin{itemize}
    \item $x = 0$: $y = -\frac{3}{\sqrt{2}}$, distance squared $= \frac{9}{2}$,
    distance $= \frac{3}{\sqrt{2}}$.
    \item $x = \pm\sqrt{2}$: $y = \frac{1}{\sqrt{2}}(2-3) = -\frac{1}{\sqrt{2}}$,
    distance squared $= 2 + \frac{1}{2} = \frac{5}{2}$, distance $= \sqrt{\frac{5}{2}}$.
\end{itemize}

The minimum distance is $\sqrt{5/2} = \sqrt{5}/\sqrt{2}$, so $a = 5$,
$b = 2$, and $a + b = 7$.
}
\end{tcolorbox}

\begin{tcolorbox}[
    prismsolutionbox,
    title=\textbf{\ours{} Solution --- Inverse Substitution (in $y$)},
]
\small{
The distance from $(x_0, y_0)$ to the origin is $\sqrt{x_0^2 + y_0^2}$,
so I need to minimize $\sqrt{x^2 + y^2}$ subject to $y =
\frac{1}{\sqrt{2}}(x^2 - 3)$.

Rather than substituting $y$ into the distance formula, I solve the
curve equation for $x^2$: $x^2 = \sqrt{2}\, y + 3$. Substituting back
into the distance formula, the distance squared becomes
$$ D^2 = x^2 + y^2 = (\sqrt{2}\, y + 3) + y^2 = y^2 + \sqrt{2}\, y + 3. $$

This is now a quadratic in $y$. Minimizing $f(y) = y^2 + \sqrt{2}\, y + 3$:
$$ f'(y) = 2y + \sqrt{2} = 0 \implies y = -\frac{\sqrt{2}}{2}. $$
Since the coefficient of $y^2$ is positive, this is a minimum. Evaluating:
$$ f\left(-\frac{\sqrt{2}}{2}\right) = \frac{1}{2} - 1 + 3 = \frac{5}{2}. $$

So $D^2 = \frac{5}{2}$, giving $D = \sqrt{5/2} = \sqrt{5}/\sqrt{2}$.
Therefore $a = 5$, $b = 2$, and $a + b = 7$.
}
\end{tcolorbox}

\begin{tcolorbox}[
    prismsolutionbox,
    title=\textbf{\ours{} Solution --- Variable Transformation ($t = x^2$)},
]
\small{
The distance from a point on the graph to the origin is
$\sqrt{x^2 + y^2}$. Substituting $y = \frac{1}{\sqrt{2}}(x^2 - 3)$:
$$ \sqrt{x^2 + \frac{(x^2 - 3)^2}{2}}. $$

To simplify, let $t = x^2$ (with $t \geq 0$). The expression becomes
$\sqrt{t + (t - 3)^2 / 2}$. Squaring and expanding:
$$ g(t) = t + \frac{(t-3)^2}{2} = t + \frac{t^2 - 6t + 9}{2}
       = \frac{t^2 - 4t + 9}{2}. $$

Minimizing $g(t)$: $g'(t) = \frac{2t - 4}{2} = t - 2$, so $t = 2$ is
the critical point. Since $g(t)$ is a quadratic opening upward, $t = 2$
is the minimum. Evaluating:
$$ g(2) = \frac{4 - 8 + 9}{2} = \frac{5}{2}. $$

The minimum distance is $\sqrt{5/2} = \sqrt{5}/\sqrt{2}$, so $a = 5$,
$b = 2$, and $a + b = 7$.
}
\end{tcolorbox}

\begin{tcolorbox}[
    prismsolutionbox,
    title=\textbf{\ours{} Solution --- Vertex-based Heuristic},
]
\small{
The function is a quadratic of the form $y = \frac{1}{\sqrt{2}} x^2 -
\frac{3}{\sqrt{2}}$. Since the coefficient of $x^2$ is positive, the
parabola opens upwards. The vertex of the parabola is the point where
the function reaches its minimum value, so the closest point on the
graph to the origin must be the vertex itself.

The vertex occurs at $x = -\frac{b}{2a} = 0$. Substituting $x = 0$
gives $y = -\frac{3}{\sqrt{2}}$. Therefore the vertex is at
$(0, -\frac{3}{\sqrt{2}})$.

The distance from the origin to the vertex is
$$ \sqrt{0^2 + \left( -\frac{3}{\sqrt{2}} \right)^2}
 = \sqrt{\frac{9}{2}} = \frac{3}{\sqrt{2}}. $$

Expressed as $\sqrt{a}/b$: $\frac{3}{\sqrt{2}} = \frac{\sqrt{9}}{\sqrt{2}}$,
so $a = 9$, $b = 2$, and $a + b = 11$.
}
\end{tcolorbox}

%% file: case_study/code_baseline.tex
\begin{tcolorbox}[
    prismsolutionbox,
    title=\textbf{Baseline Solution 1 (SFT): Stirling-Number Table},
]
\begin{lstlisting}[language=Python]
def bell_Number(n):
    # Create a table to store results of subproblems
    stirling = [[0] * (n + 1) for _ in range(n + 1)]
    # Base case
    stirling[0][0] = 1
    # Fill the table using the recurrence for Stirling numbers
    for i in range(1, n + 1):
        for j in range(1, i + 1):
            stirling[i][j] = j * stirling[i - 1][j] + stirling[i - 1][j - 1]
    # The n-th Bell number is the sum of the n-th row
    return sum(stirling[n])
\end{lstlisting}
\end{tcolorbox}

%% file: case_study/code_prism.tex

\begin{tcolorbox}[
    prismsolutionbox,
    title=\textbf{Prism Solution 1: Dynamic Programming},
]
\begin{lstlisting}[language=Python]
def bell_Number(n):
    if n == 0:
        return 1
    bell = [0] * (n + 1)
    bell[0] = 1
    for i in range(1, n + 1):
        for j in range(i):
            bell[i] += bell[j] * binomial_coefficient(i - 1, j)
    return bell[n]

def binomial_coefficient(n, k):
    if k > n - k:
        k = n - k
    c = 1
    for i in range(k):
        c = c * (n - i) // (i + 1)
    return c
\end{lstlisting}
\end{tcolorbox}

\begin{tcolorbox}[
    prismsolutionbox,
    title=\textbf{Prism Solution 2: Memoized Recursion},
]
\begin{lstlisting}[language=Python]
def bell_Number(n, memo={}):
    if n in memo:
        return memo[n]
    if n == 0 or n == 1:
        return 1
    result = 0
    for k in range(n):
        result += binomial_coefficient(n - 1, k) * bell_Number(k, memo)
    memo[n] = result
    return result

def binomial_coefficient(n, k):
    if k > n - k:
        k = n - k
    result = 1
    for i in range(k):
        result *= (n - i)
        result //= (i + 1)
    return result
\end{lstlisting}
\end{tcolorbox}

\begin{tcolorbox}[
    prismsolutionbox,
    title=\textbf{Prism Solution 3: Combinatorial Summation},
]
\begin{lstlisting}[language=Python]
def bell_Number(n):
    if n == 0 or n == 1:
        return 1
    bell = [1, 1]
    for i in range(2, n + 1):
        bell.append(sum(bell[j] * binomial_coefficient(i - 1, j) for j in range(i)))
    return bell[n]

def binomial_coefficient(n, k):
    if k > n - k:
        k = n - k
    c = 1
    for i in range(k):
        c = c * (n - i) // (i + 1)
    return c
\end{lstlisting}
\end{tcolorbox}